%% file: paper.tex
\documentclass[sigconf,natbib,authorversion,nonacm]{acmart}

\usepackage{hyperref}       
\usepackage{url}            
\usepackage{booktabs}       
\usepackage{amsfonts}       
\usepackage{nicefrac}       
\usepackage{booktabs}       
\usepackage{nicefrac}       
\usepackage{graphicx}       
\usepackage{bm}
\usepackage{dsfont}
\usepackage{natbib}

\usepackage{xcolor}
\usepackage[ruled,vlined]{algorithm2e}

\SetCommentSty{mycommfont}

\usepackage{appendix}
\usepackage{amsmath}

\newcommand{\argmax}{\mathop{\mathrm{arg\,max}}}

\AtBeginDocument{%
  \providecommand\BibTeX{{%
    \normalfont B\kern-0.5em{\scshape i\kern-0.25em b}\kern-0.8em\TeX}}}

\setcopyright{acmcopyright}
\copyrightyear{2021}
\acmYear{2021}
\acmDOI{10.1145/nnnnnnn-nnnnnnn}

\acmConference[Woodstock '18]{Woodstock '18: ACM Symposium on Neural
  Gaze Detection}{June 03--05, 2018}{Woodstock, NY}
\acmBooktitle{Woodstock '18: ACM Symposium on Neural Gaze Detection,
  June 03--05, 2018, Woodstock, NY}
\acmPrice{15.00}
\acmISBN{978-1-4503-XXXX-X/18/06}

\begin{document}

\title{Model-agnostic and Scalable Counterfactual Explanations via Reinforcement Learning}


\author{Robert-Florian Samoilescu}
\authornote{Work done during an internship at Seldon.}
\affiliation{%
  \institution{Seldon Technologies}
  \city{London}
  \country{UK}}
\email{rfs@seldon.io}

\author{Arnaud Van Looveren}
\affiliation{%
  \institution{Seldon Technologies}
  \city{London}
  \country{UK}}
\email{avl@seldon.io}

\author{Janis Klaise}
\affiliation{%
  \institution{Seldon Technologies}
  \city{London}
  \country{UK}}
\email{jk@seldon.io}


\input{sections/00-abstract}





\maketitle

\input{sections/01-introduction}

\input{sections/02-related_work}

\input{sections/03-reinforcement_learning_background}
\input{sections/04-methods}

\input{sections/05-experiments}

\input{sections/06-conclusion}

\bibliographystyle{ACM-Reference-Format}
\bibliography{references}

\onecolumn
\appendix
\appendixpage
\input{sm_sections/00-tabular_classifier}
\input{sm_sections/01-tabular_autoencoder}
\input{sm_sections/02-image_autoencoder}

\input{sm_sections/03-ddpg}

\input{sm_sections/04-mmd_covertype}
\input{sm_sections/05-hyperparameters}
\input{sm_sections/06-samples_adult}

\input{sm_sections/07-samples_mnist}
\input{sm_sections/08-samples_celeba}


\end{document}

%% file: sections/00-abstract.tex
\begin{abstract}

Counterfactual instances are a powerful tool to obtain valuable insights into automated decision processes, describing the necessary minimal changes in the input space to alter the prediction towards a desired target. Most previous approaches require a separate, computationally expensive optimization procedure per instance, making them impractical for both large amounts of data and high-dimensional data. Moreover, these methods are often restricted to certain subclasses of machine learning models (e.g. differentiable or tree-based models). In this work, we propose a deep reinforcement learning approach that transforms the optimization procedure into an end-to-end learnable process, allowing us to generate batches of counterfactual instances in a single forward pass. Our experiments on real-world data show that our method i) is model-agnostic (does not assume differentiability), relying only on feedback from model predictions; ii) allows for generating target-conditional counterfactual instances; iii) allows for flexible feature range constraints for numerical and categorical attributes, including the immutability of protected features (e.g. gender, race); iv) is easily extended to other data modalities such as images.

\end{abstract}

%% file: sections/01-introduction.tex
\section{Introduction}
\label{sec:introduction}

Recent advances in machine learning algorithms have resulted in large-scale adoption of predictive models across industries. However, the black-box nature of machine learning systems makes it difficult to build trust in algorithmic decision making, especially in sensitive and safety critical areas where humans are directly affected. For example, using machine learning as part of the decision making process for loan approvals, university admissions or employment applications, simple rejection feedback can be misinterpreted or raise serious concerns regarding the ability of an institution to provide equal opportunities to all applicants.


\emph{Counterfactual instances} (a.k.a. \emph{counterfactual explanations, counterfactuals})~\citep{wachter2017counterfactual,karimi2020survey,stepin2021survey} are a powerful tool to obtain insight into the underlying decision process exhibited by a black-box model, describing the necessary minimal changes in the input space to alter the prediction towards a desired target. To be of practical use, a counterfactual should be \emph{sparse}---close (using some distance measure) to the original instance---and indistinguishable from real instances, that is, it should be \emph{in-distribution}. Thus, for a loan application system that currently outputs a rejection for a given individual, a counterfactual explanation should suggest plausible minimal changes in the feature values that the applicant could perform to get the loan accepted leading to actionable recourse~\citep{joshi2019towards}.

\begin{figure}[!htbp]
    \centering
    \includegraphics[width=0.96\linewidth]{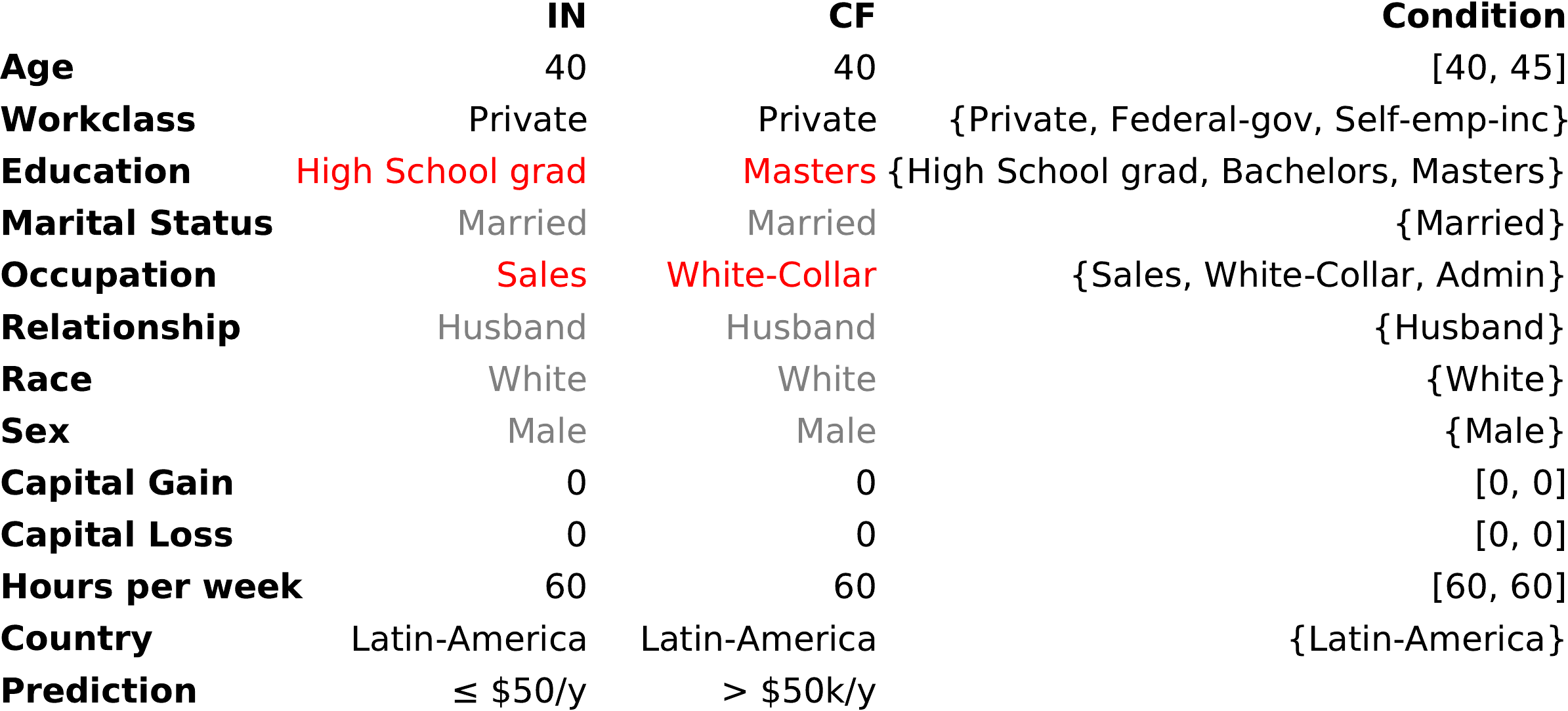}
    \caption{Conditional counterfactual instance on Adult dataset. IN---original instance, CF---counterfactual instance, Condition---feature range/subset constraints. Grayed out feature values correspond to immutable features. Highlighted in red, the feature changes required to alter the prediction of a black-box model (here from $\leq\$50k/y$ to $>\$50k/y$). See Appendix~\ref{app:adult} for more examples.}
    \label{fig:cf_example}
\end{figure}

A desirable property of a method for generating counterfactuals is to allow feature conditioning. Real-world datasets usually include immutable features such as gender or race, which should remain unchanged throughout the counterfactual search procedure. A natural extension of immutability is to restrict a feature to a subset or an interval of values. Thus, following the same loan application example, a customer might be willing to improve their education level from a \textit{High-school graduate} to \textit{Bachelor's} or \textit{Master's}, but not further.  Similarly, a numerical feature such as \textit{Age} should only increase for a counterfactual to be actionable. To enable such feature conditioning, we propose to use a conditioning vector to guide the generation process. An example of this scenario on the Adult dataset \cite{Dua:2019} is shown in Figure~\ref{fig:cf_example} where we enforce feature conditioning and immutability. Our method successfully flips the classification label by suggesting a change in the \textit{Education} level, from \textit{High School grad} to \textit{Master's}, and in the \textit{Occupation}, from \textit{Sales} to \textit{White-Collar}, producing an actionable counterfactual explanation with respect to the user-specified feature constraints.


Previous approaches to finding counterfactual instances have focused primarily on iterative procedures by minimizing an objective function that favors sparse and in-distribution results. Some methods are only suitable for differentiable models since they require access to model gradients.
Moreover, most current methods require a separate, computationally expensive optimization procedure per instance, making them impractical for both large amounts of data and high-dimensional data.


\paragraph{Our contributions.}
We propose a deep reinforcement learning approach that transforms the optimization procedure into a learnable process, allowing us to generate batches of counterfactual instances in a single forward pass. Our training pipeline is model-agnostic and relies only on prediction feedback by querying the black-box model.
Furthermore, our method allows target and feature conditioning, and is applicable to a variety of tasks such as multi-class classification and regression.

We focus primarily on the tabular data setting and evaluate our method on multiple types of models and datasets, for which we obtain competitive results compared to the existing algorithms. Additionally, we show that our approach is flexible and easily extendable to other data modalities such as images.

In summary, our contributions are as follows:
\begin{itemize}
    \item Model-agnostic, target-conditional framework primarily focused on heterogeneous tabular datasets.
    \item Flexible feature range constraints for numerical and categorical features.
    \item Fast counterfactual generation process since no iterative optimization procedure is required when producing counterfactual instances.
    \item Easily extendable framework to other data modalities.
\end{itemize}

%% file: sections/02-related_work.tex
\section{Related work}
\label{sec:related_work}
Counterfactual explanation methods have recently seen an explosive growth in interest by the academic community. We refer the reader to the recent surveys of \citet{karimi2020survey,stepin2021survey} for a more comprehensive review of the area.

Most counterfactual explanation methods perturb the original instance under proximity constraints until the desired model prediction is achieved~\cite{wachter2017counterfactual, mothilal2020explaining, cheng2020dece} or conduct a heuristic search~\cite{martens_doc,laugel2017inverse}. These approaches require a separate optimization process for each explained instance and often result in out-of-distribution counterfactuals when the perturbations are applied in the (high-dimensional) input space. Attempts to generate more realistic, in-distribution counterfactuals include the addition of auxiliary losses such as the counterfactual's reconstruction error using a pre-trained autoencoder \cite{dhurandhar2018explanations} or by guiding the perturbations towards class-specific prototypes \citep{van2019interpretable}. Other methods leverage pre-trained generative models such as conditional GANs \cite{mirza_cgan, liu_cfintro} or variational autoencoders \cite{joshi2019towards} to improve the realism of the proposed counterfactuals. \citet{joshi2019towards} apply perturbations in the latent space of a variational autoencoder to bypass the more challenging optimization process in the input feature space. 

Many of the above methods rely on gradient-based optimization which restricts the application to differentiable models. In practice we are often only able to query the model, thus we are interested in the model-agnostic, black-box setting. This is especially relevant for (mixed-type) tabular data where non-differentiable models such as Random Forests \cite{randomforest} or XGBoost \cite{xgboost} remain very popular. LORE \cite{guidotti2018local} is a model agnostic method which, similar to LIME \cite{ribeiro2016should}, learns a local interpretable surrogate model around the instance of interest. The surrogate model is trained on a set of synthetic instances created by a genetic algorithm in the neighbourhood of the explained instance. The local model provides a set of decision rules which change the original model prediction and generate a counterfactual. Similarly, \citet{white2019measurable} extract local rules from decision trees to generate counterfactual explanations \cite{dhurandhar2018explanations}. DiCE \cite{mothilal2020explaining} aims to generate a diverse set of black box counterfactual explanations by using determinantal point processes \cite{determinantalpointprocesses} under both proximity and feature level user constraints. The search process can be done at random or via a genetic algorithm. \citet{hashemi2020permuteattack} follow the same paradigm and refine the sampling procedure to favour in-distribution counterfactuals. \citet{sharma2019certifai} also use evolutionary strategies which allow for feature conditioning to obtain counterfactual explanations. Minimum Observable (MO) \cite{wexler2019if} on the other hand simply returns the closest instance in the training set which belongs to the target class. The distance is measured by a combination of the $\mathcal{L}_1$ and $\mathcal{L}_0$ norms of respectively the standardized numerical and categorical features. This requires the training set and model predictions to be kept in memory.

The methods mentioned so far either need access to a training set when generating the explanation or rely on a time consuming iterative search procedure for every explained instance. This bottleneck can be resolved by using class-conditional generative models which are able to generate batches of sparse, in-distribution counterfactual instances in a single forward pass for various data modalities such as images, time series or tabular data \cite{van2021conditional, oh_bornid, mahajan_preserving}. However, end-to-end training of the counterfactual generator requires backpropagation of the gradients through the model, limiting the applicability to differentiable models.

We propose a model-agnostic counterfactual generator able to generate batches of explanations with a single prediction. The generator is trained using reinforcement learning (RL) and is able to learn solely from sparse model prediction rewards. Our method allows for flexible feature conditioning and can easily be extended to various data modalities.

%% file: sections/03-reinforcement_learning_background.tex
\section{Reinforcement learning background}
\label{sec:reinforcement_learning_background}

In this section, we provide a brief overview of the reinforcement learning (RL) paradigm and domain-specific terminology. We consider a standard model-free RL problem where an agent learns an optimal behavior policy by repeated interactions with the environment, intending to maximize the expected cumulative reward received from the environment. One approach to learn an optimal policy is to approximate the Q-function, which estimates the reward that an agent will obtain by applying a particular action in a given state. Thus, if we know the optimal action-value function $Q^{*}(s, a)$, in a state $s$ and for any action $a$, then the optimal policy is given by:
$$
	a^{*}(s) = \argmax_{a}{Q^{*}(s, a)}.
$$

For a discrete action space, the maximization procedure requires the evaluation of the state-action pair for the available actions and retrieving the action that maximizes the Q function for a given state. On the other hand, for a continuous action space, an exhaustive search is impossible and alternatively maximizing the $Q$ function through an iterative procedure can be a computational bottleneck. Deep Deterministic Policy Gradient (DDPG) \cite{lillicrap2015continuous} addresses those issues by interleaving a state-action function approximator $Q$ (the critic) of $Q^{*}(s, a)$ with learning an approximator $\mu$ (the actor) for the optimal action $a^{*}(s)$. The method assumes that the critic is differentiable with respect to the action argument thus allowing to optimize the actor's parameters efficiently through gradient-based methods. DDPG approximates the solution of the expensive maximization procedure through a learnable process:

\begin{equation} \label{eq:critic}
    \mu(s) \approx \argmax_{a}{Q(s, a)}.
\end{equation}



Similar to other deep Q-learning algorithms \cite{mnih2013playing}, DDPG uses a replay buffer for sample efficiency. Instead of immediately discarding the current experience, the method stores it in the replay buffer for later use. Each training phase consists of sampling a batch of experiences uniformly and taking a gradient step. Updates based on old experience prevent overfitting and increase training stability.

%% file: sections/04-methods.tex
\section{Method}
\label{sec:method}

\input{subsections/03-methods/00-problem_statement}
\input{subsections/03-methods/01-target_conditioning}
\input{subsections/03-methods/02-attribute_conditioning}
\input{subsections/03-methods/03-training_procedure}
\input{subsections/03-methods/04-handling_heterogeneous_data}
\input{subsections/03-methods/05-inference}

%% file: subsections/03-methods/00-problem_statement.tex
\subsection{Problem statement}
\label{subsec:method_problem_statement}

A counterfactual explanation of a given instance represents a sparse, in-distribution example that alters the model prediction towards a specified target. Following the notation of \citet{van2021conditional}, let $x$ be the original instance, $M$ a black-box model, $y_M = M(x)$ the model prediction on $x$ and $y_T$ the target prediction. Our goal is to produce a counterfactual instance $x_{CF} = x + \delta_{CF}$ where $\delta_{CF}$ represents a sparse perturbation vector such that $y_T = M(x_{CF})$. Instead of solving an optimization problem for each input instance, we train a generative model which models the counterfactual instances $x_{CF}$ directly and allows for feature level constraints via an optional conditioning vector $c$. A conditional counterfactual explanation $x_{CF}$ therefore depends on the tuple $s=(x,y_M,y_T,c)$.




Since we don't assume the model $M$ to be differentiable we train the counterfactual generator using reinforcement learning as discussed in Section~\ref{sec:reinforcement_learning_background} with $s$ representing the state and the actor network $\mu(s)$ taking the role of the counterfactual generator. This model-agnostic training pipeline is compatible with various data modalities and only uses sparse model prediction feedback as a reward. For a classification model returning the predicted class label the reward can be defined by an indicator function, $R=\mathds{1}(M(x_{CF}) = y_T)$. The reward for a regression model, on the other hand, is proportional to the proximity of $M(x_{CF})$ to the regression target $y_T$. 


Instead of directly modelling the perturbation vector $\delta_{CF}$ in the potentially high-dimensional input space, we first train an autoencoder. The weights of the encoder are frozen and $\mu$ applies the counterfactual perturbations in the latent space of the encoder. The pre-trained decoder maps the counterfactual embedding back to the input feature space. Since $\mu$ operates in the continuous latent space we use the sample efficient DDPG \cite{lillicrap2015continuous} method. For the remainder of the paper, we denote by $enc$ and $dec$ the encoder and the decoder networks, respectively.

%% file: subsections/03-methods/01-target_conditioning.tex
\subsection{Target conditioning}
\label{subsec:method_target_conditioning}


The counterfactual generation process generalizes to a variety of tasks and target types such as multi-class classification and regression. For classification tasks, we condition on a one-hot encoded representation of the target prediction $y_T$ as well as on the predicted label $y_M$ on the original instance $x$. 
While conditioning solely on the class labels represents the most generic use case since it does not require the model output to be probabilistic (e.g. SVM's \cite{svm}), we could also condition the counterfactual generator on soft prediction targets such as the probability of the target class or the full prediction distribution over all the classes.
Note that our approach does not require the ground truth and relies only on querying the black-box model. During training, we sample the target prediction $y_T$ uniformly at random from all possible labels, including $y_T = y_M$ for the identity mapping.



%% file: subsections/03-methods/02-attribute_conditioning.tex
\subsection{Feature conditioning}
\label{subsec:method_attribute_conditioning}

In many real-world applications, some of the input features are immutable, have restricted feature ranges or are constrained to a subset of all possible feature values. These constraints need to be taken into account when generating actionable counterfactual instances. For instance \textit{age} and \textit{marital status} could be features in the loan application example. An actionable counterfactual should however only be able to increase the numerical \textit{age} feature and keep the categorical \textit{marital status} feature unchanged. To achieve this we condition the counterfactual generator on a conditioning vector $c$. $c$ is constructed by concatenating per-feature conditioning vectors, defined as follows:


\begin{itemize}
    
    \item For a numerical feature with a minimum value $a_{min}$ and a maximum value $a_{max}$, we append to $c$ the values $-p_{min}$, $p_{max}$, where $p_{min}, p_{max} \in [0, 1]$. The range $[-p_{min}$, $p_{max}]$ encodes a shift and scale-invariant representation of the interval $[a - p_{min} (a_{max} - a_{min}), a + p_{max} (a_{max} - a_{min})]$, where $a$ is the original feature value.
    During training, $p_{min}$ and $p_{max}$ are sampled $\sim Beta(2, 2)$ for each unconstrained feature. Immutable features correspond to setting $p_{min} = p_{max} = 0$. Features allowed to only increase or decrease correspond to setting $p_{min} = 0$ or $p_{max} = 0$, respectively. Following the example in Figure \ref{fig:cf_example}, allowing the \emph{age} feature to increase by up to $5$ years is encoded in the conditional vector $c$ by taking $p_{min}=0,p_{max}=0.1$, assuming a minimum age of 10 and a maximum age of 60 years in the training set ($5 = 0.1 \cdot (60 - 10)$).
    
    
    \item For a categorical feature of cardinality $K$ we condition the subset of allowed feature values through a binary mask of dimension $K$. When training the counterfactual generator, the mask values are sampled $\sim Bern(0.5)$. For immutable features, only the original input feature value is set to one in the binary mask. Following the example in Figure \ref{fig:cf_example}, the immutability of the \emph{marital status} having the current value \emph{married} is encoded through the binary sequence $(1, 0, 0)$, given an ordering of the possible feature values \{\emph{married}, \emph{unmarried}, \emph{divorced}\}. 
\end{itemize}


Following the decoding phase, as part of post-processing (denoted by a function $pp$), the numerical values are clipped within the desired range, and categorical values are conditionally sampled according to their masking vector. This step ensures that the generated counterfactual respects the desired feature conditioning before passing it to the model. Note that our method is flexible and allows non-differentiable post-processing such as casting features to their original data types (e.g. converting a decoded floating point \emph{age} to an integer: 40 = int(40.3)) and categorical mapping (e.g., \emph{marital status} distribution/one-hot encoding to the \emph{married} value) since we rely solely on the sparse model prediction reward.



%% file: subsections/03-methods/03-training_procedure.tex
\subsection{Training procedure}
\label{subsec:method_training_procedure}

The DDPG algorithm requires two separate networks, an actor $\mu$ and a critic $Q$. Given the encoded representation of the input instance $z=enc(x)$, the model prediction $y_M$, the target prediction $y_T$ and the conditioning vector $c$, the actor outputs the counterfactual's latent representation $z_{CF}=\mu(z,y_M,y_T,c)$. The decoder then projects the embedding $z_{CF}$ back to the original input space, followed by optional post-processing.


We restrict the components of the embedding representation to $[-1, +1]$ through a \emph{tanh} non-linearity. During the first $N$ steps the latent components are sampled uniformly from $[-1, +1]$ to encourage exploration. Afterwards, Gaussian noise $\epsilon\sim \mathcal{N}(0, 0.1)$ is added to the actor's output. The experience is stored in a replay buffer from which we uniformly sample during the training phase.


The training step consists of simultaneously optimizing the actor and critic networks. The critic regresses on the reward $R$ determined by the model prediction, while the actor maximizes the critic's output for the given instance according to Equation \ref{eq:critic}. The actor also minimizes two objectives to encourage the generation of sparse, in-distribution counterfactuals. The sparsity loss $\mathcal{L}_{\textit{sparsity}}$ operates on the decoded counterfactual $x_{CF}$ and combines the $\mathcal{L}_{1}$ loss over the standardized numerical features and the $\mathcal{L}_{0}$ loss over the categorical ones. The consistency loss $\mathcal{L}_{\textit{consist}}$ \cite{cyclegan} aims to encode the counterfactual $x_{CF}$ back to the same latent representation where it was decoded from and helps to produce in-distribution counterfactual instances. We formally define $\mathcal{L}_{\textit{consist}}=\Vert z_{CF} - enc(pp(x_{CF}, c))\Vert_{2}^{2}$.


Notice that our setup is equivalent to a Markov decision process with a one-step horizon, which does not require bootstrapping to compute the critic's target, increasing stability and simplifying the training pipeline. We provide a full description of the training procedure in Algorithm \ref{algo:ddpg}. See also Appendix~\ref{app:ddpg} for details of the actor and critic architectures.

\begin{algorithm}[!htbp]
\caption{Training procedure}
\label{algo:ddpg}
\SetAlgoLined
\SetKwInOut{Input}{Input}
\SetKwInOut{Output}{Output}
\Input{$M$ - black-box model, loss hyperparameters $\lambda_s,\lambda_c$}
\Output{$\mu$ - trained actor network used for counterfactual generation.}

Load pre-trained encoder $enc$, and pre-trained decoder $dec$.

Randomly initialize the actor $\mu(\cdot;\theta_{\mu})$ and the critic $Q(\cdot;\theta_{Q})$.\\

Initialize the replay buffer $\mathcal{D}$.\\

Define reward function $f(\cdot,\cdot)$.\\ 

Define post-processing function $pp$. \\

\For{however many steps}{
    Sample batch of input data $x$. \\
    
    Construct random target $y_T$ and conditioning vector $c$. \\
    
    Compute $y_M = M(x)$, $z = enc(x)$, $z_{CF}=\mu(z, y_M, y_T, c; \theta_{\mu})$.\\

    Select $\tilde{z}_{CF} = clip(z_{CF} + \epsilon, -1, 1)$, $\epsilon \sim \mathcal{N}(0, 0.1)$.\\
    
    Decode $\tilde{x}_{CF} = pp(dec(\tilde{z}_{CF}), c)$.\\
    
    
    Observe $R=f(M(\tilde{x}_{CF}), y_T)$.  

    Store $(x, z, y_M, y_T, c, \tilde{z}_{CF}, R)$ in the replay buffer $\mathcal{D}$.\\
    
    \If{time to update}{
        \For{however many updates} {
            Sample uniformly a batch of $B$ experiences $\mathcal{B} = \{(z, y_M, y_T, c, \tilde{z}_{CF}, R)\}$ from $\mathcal{D}$. \\
            
            Update critic by one-step gradient descent using 
            \begin{flalign*}
                \quad &\nabla_{\theta_{Q}}\frac{1}{|B|} \sum_{B}{(Q(z, y_M, y_T, c, \tilde{z}_{CF}) - R)^2} &\\
            \end{flalign*}

            Compute $z_{CF} = \mu(z, y_M, y_T, c; \theta_{\mu})$, $x_{CF} = dec(z_{CF})$, 
            \begin{flalign*}
                \mathcal{L}_{max} &= -\frac{1}{|B|} \sum_{B}{Q(z, y_M, y_T, c, z_{CF})}, &\\
                \mathcal{L}_{sparsity} &= \frac{1}{|B|} \sum_{B}{[\mathcal{L}_{1}(x, x_{CF}) + \mathcal{L}_{0}(x, x_{CF})]}, \\
                \mathcal{L}_{consist} &= \frac{1}{|B|} \sum_{B}{(enc(pp(x_{CF}, c)) - z_{CF})^{2}}.
            \end{flalign*}
            
                    
            
            
            Update actor by one-step gradient descent using
            \begin{flalign*}
                \quad &\nabla_{\theta_{\mu}} (\mathcal{L}_{max} + \lambda_{s} \mathcal{L}_{sparsity} + \lambda_{c} \mathcal{L}_{consist}) &\\
            \end{flalign*}
        }
    }
}
\end{algorithm}

%% file: subsections/03-methods/04-handling_heterogeneous_data.tex
\subsection{Handling heterogeneous data types}
\label{subsec:handling_heterogeneous_datatypes}

Our method is versatile and easily adaptable to multiple data modalities by changing only the autoencoder component. As opposed to homogeneous data such as images where the observed features share the same likelihood function, tabular data require more flexibility to capture relations across heterogeneous data types. We therefore define the decoder as a multi-head, fully connected network for which we model each feature feature individually as follows:

\begin{itemize}
    \item For a real-valued feature $d$ of an instance $x$ with a corresponding encoding $z$, we use a Gaussian likelihood model with constant variance given by: 
$$
p(x_{d} | z) = \mathcal{N}(x_{d} | h_{d}(z), \sigma_{d})
$$
where $h_{d}$ is the decoder head.
    \item We apply one-hot encoding to categorical features and model feature values using the categorical distribution. Following the same notation from above, the probability for a category $k$ is given by:
$$
p(x_{d} = k | z) = \frac{exp(-h_{dk}(z))}{\sum_{i=1}^{K}{exp(-h_{di}(z)}}
$$
where $h$ is the decoder head, and $K$ is the total number of feature values. During training and generation, we pick the most probable feature according to the output distribution.
\end{itemize}

%% file: subsections/03-methods/05-inference.tex
\subsection{Generating explanations}
\label{subsec:method_inference}

Our counterfactual generation method does not require an iterative optimization procedure per input instance and allows us to generate batches of conditional counterfactual explanations in a single forward pass. This allows the method to scale to large and high-dimensional datasets. The generation step requires a user-specified target $y_T$, an optional feature-conditioning tensor $c$, and the black-box model $M$ to label the given input. The pre-trained encoder projects the test instance onto the latent space where the actor-network generates the embedding representation of the counterfactual, given the target and feature-conditioning vector. 

In the final phase, we decode the counterfactual embedding followed by an optional post-processing step which ensures that the constraints are respected, casts features to their desired types and maps categorical variables to the original input space. We provide a full description of the generation procedure in Algorithm \ref{algo:inference}.


\begin{algorithm}[!htbp]
\caption{Generating explanations}
\label{algo:inference}
\SetAlgoLined

\SetKwInOut{Input}{Input}
\SetKwInOut{Result}{Result}
\Input{$x$ - original instance, $y_T$ - prediction target, $c$ - optional feature-conditioning vector, $M$ - black-box model.}
\Result{$x_{CF}$ - counterfactual instance.}

Load pre-trained actor $\mu$, encoder $enc$, decoder $dec$, and post-processing function $pp$.\\

Compute $z = enc(x)$ input encoding representation.\\

Compute $y_M = M(x)$ model's prediction.\\

Generate $z_{CF} = \mu(z, y_M, y_T, c)$ counterfactual embedding.\\

Decode and post-process $\tilde{x}_{CF} = pp(dec(z_{CF}), c)$ counterfactual instance.\\
\end{algorithm}

%% file: sections/05-experiments.tex
\section{Experiments}
\label{sec:experiments}

\input{subsections/04-experiments/008-classifier_accuracy_table}
\input{subsections/04-experiments/009-validity_table}
\input{subsections/04-experiments/00-setup}
\input{subsections/04-experiments/01-validity}

\input{subsections/04-experiments/019-sparsity_tables}
\input{subsections/04-experiments/02-sparsity}

\input{subsections/04-experiments/039-umap_figure}
\input{subsections/04-experiments/029-in_distribution_tables}
\input{subsections/04-experiments/049-diversity_figure}
\input{subsections/04-experiments/051-flexibility_figure}

\input{subsections/04-experiments/03-in_distribution}

\input{subsections/04-experiments/04-diversity}

\input{subsections/04-experiments/05-flexibility}

%% file: subsections/04-experiments/008-classifier_accuracy_table.tex
\begin{table*}[!htbp]
  \small
  \caption{Train/test classification accuracy per model and dataset.}
  \label{tab:classifiers_accuracy}
  \centering
  
  \begin{tabular}{lccccc}
    \toprule
    &               \multicolumn{5}{c}{\textbf{Accuracy train/test}(\%)} \\
                    \cmidrule(r){2-6} 
    \textbf{Model}          &\textbf{Adult}     &\textbf{Breast cancer}     &\textbf{Covertype}     &\textbf{Portuguese Bank}       &\textbf{Spambase}      \\
    \midrule
    LR                      &85 / 85        & 97 / 96               &94 / 93            &85 / 85                    &61 / 60          \\
    DT	                    &87 / 85        & 95 / 95               &97 / 91            &98 / 86                    &98 / 82          \\
    RF          	        &88 / 86        & 97 / 96               &99 / 95            &100 / 91                   &100 / 88          \\
    LSCV        	        &85 / 85        & 98 / 96               &94 / 93            &87 / 87                    &68 / 68           \\
    XGBC                    &88 / 87        & 100 / 96              &100 / 95           &90 / 90                    &100 / 91          \\
    \bottomrule
  \end{tabular}
\end{table*}

%% file: subsections/04-experiments/009-validity_table.tex
\begin{table*}[!htbp]
  \small
  \caption{Counterfactual validity---percentage of generated counterfactuals of the desired target label (mean$\pm$std over 5 classifiers and 3 random seeds).}
  \label{tab:counterfactuals_accuracy}
  \centering
  
  \begin{tabular}{lllllll}
    \toprule
    &               \multicolumn{5}{c}{\textbf{Validity} (\%)} \\
                    \cmidrule(r){2-6} 
    \textbf{Method}                 &\textbf{Adult}             &\textbf{Breast cancer}            &\textbf{Covertype}              &\textbf{Portuguese Bank}              &\textbf{Spambase}   \\
    \midrule
    LORE                            &$18.08\pm5.27$              &$25.95\pm34.33$	               &$15.19\pm7.75$                  &$19.07\pm9.75$                         &$9.53\pm5.91$      \\
    MO	                            &$91.00\pm1.12$              &$100.00\pm0.00$                 &$100.00\pm0.00$                &$100.00\pm0.00$	                    &$100.00\pm0.00$   \\
    DiCE (random)	                &$99.93\pm0.16$	             &$100.00\pm0.00$	               &$92.67\pm11.04$	                &$99.98\pm0.04$	                        &$99.58\pm0.60$    \\
    DiCE (genetic)	                &$33.94\pm11.89$	         &$60.86\pm14.18$	               &$72.09\pm18.40$	                &$90.97\pm8.73$	                        &$40.93\pm2.28$     \\
    DiCE (binary, random)               &$100.00\pm0.00$	         &$100.00\pm0.00$	               &$92.66\pm11.03$	                &$99.99\pm0.03$	                        &$99.87\pm0.25$   \\
    DiCE (binary, genetic)              &$51.28\pm5.00$	             &$74.57\pm6.92$	               &$77.55\pm14.49$	                &$98.50\pm1.78$	                        &$40.72\pm2.07$   \\
    Ours (w/o AE)                    &$98.59\pm0.97$	             &$91.24\pm9.96$	               &$91.11\pm6.58$	                &$99.38\pm1.24$	                        &$99.72\pm0.41$  \\
    Ours (w/o AE, diverse)             &$97.17\pm0.87$	             &$91.19\pm3.73$	               &$80.07\pm7.54$	                &$98.23\pm1.20$	                        &$99.07\pm0.99$  \\
    Ours                            &$98.59\pm0.97$              &$99.24\pm0.76$	               &$86.81\pm13.68$	                &$98.27\pm1.53$	                        &$99.18\pm0.97$     \\
    Ours (diverse)                      &$97.50\pm0.58$	             &$96.90\pm1.76$	               &$76.55\pm16.65$	                &$96.10\pm2.78$	                        &$98.67\pm1.24$     \\
    \bottomrule
  \end{tabular}
\end{table*}

%% file: subsections/04-experiments/00-setup.tex
We evaluate our method on multiple data modalities and various classifiers.  We focus primarily on the tabular setting for which we selected five datasets:

\begin{itemize}
    \item Adult Census \cite{Dua:2019}---32,561 instances with 4 numerical and 8 categorical features (we use the pre-processed version from~\cite{alibi}) along with a binary classification label. In all experiments, we allow  \textit{age} to change only in the positive direction, and consider the following features as immutable: \textit{marital status}, \textit{relationship}, \textit{race}, \textit{sex}.
    
    \item Breast Cancer \cite{Zwitter}---699 instances with 9 categorical/ordinal features along with a binary classification label.
    
    \item Covertype \cite{Blackard}---581,012 instances with 10 numerical and 2 categorical features along with a multi-classification label (7 classes). For our experiments, we only consider 10\% of the available data. Due to high-class imbalance, we apply data balancing during training by adjusting the weights inversely proportional to the class frequencies.
    
    \item Portuguese Bank \cite{moro2014data}---40,841 instances with 7 numerical and 8 categorical features along with a binary classification label. In all experiments, we allow \textit{age} to change only in the positive direction. We apply the same data balancing as with Covertype.
    
    \item Spambase \cite{Dua:2019}---4,601 instances with 57 numerical features along with a binary classification label.
\end{itemize}

We test our method on five black-box classifiers: Logistic Regression (LR), Decision Tree (DT), Linear Support Vector Classifier (LSVC), Random Forest (RF), and XGBoost Classifier (XGBC). We trained all classifiers on 80\% of the data and performed model selection via cross-validation (see Appendix~\ref{app:tab_class}). Table \ref{tab:classifiers_accuracy} shows the train and test accuracy performance of each classifier. Additionally, our counterfactual generation method requires a pre-trained autoencoder per dataset. Thus, we trained a fully connected autoencoder keeping the same train-test split (see Appendix~\ref{app:tab_ae}).


We analyze the performance of our method by including four of its variants. First, we perform an ablation study on the impact of the autoencoder by removing it from the generation pipeline since the tabular dataset is already low-dimensional (denoted \emph{w/o AE}). Then, we investigate the ability to generate diverse counterfactuals by conditioning the generation process on a randomly sampled conditional vector $c$, as described in Section \ref{subsec:method_attribute_conditioning} (denoted \emph{diverse}).  For the two pipeline settings (with and without the AE), we used the same hyperparameters across all the datasets and classifiers. The configuration with the autoencoder, uses a sparsity coefficient $\lambda_s=0.5$ and a consistency coefficient $\lambda_c=0.5$.  The configuration without the autoencoder only uses a sparsity coefficient $\lambda_s=0.1$ since the consistency loss is not defined. See Appendix~\ref{app:hyper} for hyperparameter search results.

We compare our methods with three model-agnostic, gradient-free, tabular-oriented baselines, including their variants: LORE \cite{guidotti2018local}, Minimum Observable (MO) \cite{wexler2019if}, and DiCE \cite{mothilal2020explaining}. We analyzed two variants of DiCE: the \emph{random} approach, which incrementally increases the number of features to sample from, and the \emph{genetic} approach, which follows a standard evolutionary procedure.  In addition to the two standard DiCE procedures which use soft target labels (e.g., [0.7, 0.3]), we experimented with a hard/binary representation (e.g., [1, 0]) of the output probability distribution (denoted \emph{binary}). 

We evaluate all methods on a maximum of 1000 samples (where the test dataset size permits), for which we randomly generate target labels that differ from the original classification labels. To ensure the termination of the counterfactual generation process, we restrict the running time to a window of 10 seconds chosen to exceed the average generation time across all baselines. Furthermore, the constraint of 10 seconds provides a reasonable trade-off to generate enough samples to compute meaningful statics within a maximum of 2.7 hours per run. In contrast, our method generates all counterfactuals within a few seconds. We keep the default hyper-parameters for all baselines, except for LORE, where we reduce the population size by a factor of 10 to fit the time constraints without significantly affecting the overall performance. We extend LORE to allow for immutable features (e.g., race, gender) and range constraints (e.g., strictly increasing age) by restricting the neighborhood generation function to sample feature values given a conditioning vector.


We analyze the methods using the following evaluation metrics:

\begin{itemize}
    \item \emph{validity}---the percentage of counterfactual instances classified as the intended target.
    \item \emph{sparsity}---the $\mathcal{L}_{0}$ and $\mathcal{L}_{1}$ distance for categorical and standardized numerical features, respectively.
    \item \emph{in-distributionness}---the target-conditional maximum mean discrepancy (MMD) \cite{grettonmmd} which we compute by taking the MMD between the counterfactuals of a given target class and the training instances with the same target prediction. This measures whether the generated counterfactual instances come from the same distribution as the original data for the specified target class. Note that the counterfactuals do not have to be valid.
\end{itemize}



%% file: subsections/04-experiments/01-validity.tex
\subsection{Validity}
\label{subsec:experiments_validity}

We compute the validity as the percentage of counterfactual instances classified by the black-box model as the intended target in Table \ref{tab:counterfactuals_accuracy}. Our method generates counterfactuals with high validity across multiple datasets and classifiers, competitive with DiCE (random) and MO, and significantly outperforming DiCE (genetic) and LORE. While the different variations of our method perform consistently well, including the autoencoder benefits the validity when all the features are categorical such as in the breast cancer dataset. This is not surprising since DDPG is designed to work in continuous space. Note that MO can always find a counterfactual as long as there is at least one instance in the training set which complies with the counterfactual constraints and predicted target class since it does a nearest neighbour search on the training set. However, if there is no instance in the training set which satisfies the constraints it becomes impossible to generate a valid counterfactual. We attribute the low validity score of LORE to improper local decision boundary approximation. A change in the prediction of the surrogate model does not guarantee the same desired change towards the target class for the original classification model. Likewise, DiCE (genetic) does not ensure the generation of a valid counterfactual. The evolutionary strategy objective is a weighted sum between the divergence and proximity losses, which can favor sparsity over the classification change.




%% file: subsections/04-experiments/019-sparsity_tables.tex
\begin{table*}[!htbp]
  \small
  \caption{Counterfactual sparsity---$\mathcal{L}_{0}$ distance over categorical features (mean$\pm$std over 5 classifiers and 3 random seeds).}
  \label{tab:l0}
  \centering
  
  \begin{tabular}{llllll}
    \toprule
    &               \multicolumn{5}{c}{\boldsymbol{$\mathcal{L}_{0}$} \textbf{distance}} \\
                    \cmidrule(r){2-6} 
    \textbf{Method}                 &\textbf{Adult}         &\textbf{Breast cancer}     &\textbf{Covertype}         &\textbf{Portuguese Bank}               &\textbf{Spambase} \\
    \midrule
    LORE                            &$0.09\pm0.01$            &$0.11\pm0.01$                &$0.23\pm0.11$                &$0.04\pm0.02$	                &- \\
    MO	                            &$0.20\pm0.01$            &$0.53\pm0.06$                &$0.45\pm0.04$                &$0.22\pm0.08$	                &- \\
    DiCE (random)	                &$0.05\pm0.01$	          &$0.29\pm0.05$	            &$0.22\pm0.09$	              &$0.07\pm0.010$	                &- \\
    DiCE (genetic)	                &$0.18\pm0.01$	          &$0.56\pm0.04$	            &$0.45\pm0.03$	              &$0.23\pm0.04$                    &- \\
    DiCE (binary, random)               &$0.05\pm0.01$	          &$0.29\pm0.05$	            &$0.22\pm0.09$	              &$0.07\pm0.01$	                &- \\
    DiCE (binary, genetic)              &$0.19\pm0.01$	          &$0.56\pm0.03$	            &$0.46\pm0.04$	              &$0.24\pm0.05$	                &- \\
    Ours (w/o AE)                    &$0.13\pm0.10$	          &$0.48\pm0.09$	            &$0.47\pm0.13$	              &$0.31\pm0.07$	                &- \\
    Ours (w/o AE, diverse)              &$0.11\pm0.09$	          &$0.45\pm0.10$	            &$0.34\pm0.11$	              &$0.25\pm0.08$	                &- \\
    Ours                            &$0.11\pm0.10$	          &$0.44\pm0.11$	            &$0.41\pm0.15$	              &$0.23\pm0.09$                    &- \\
    Ours (diverse)                      &$0.10\pm0.09$	          &$0.43\pm0.10$	            &$0.34\pm0.14$	              &$0.17\pm0.06$	                &- \\
    \bottomrule
  \end{tabular}
\end{table*}

\begin{table*}[!htbp]
  \small
  \caption{Counterfactual sparsity---$\mathcal{L}_{1}$ distance over standardized numerical features (mean$\pm$std over 5 classifiers and 3 random seeds).}
  \label{tab:l1}
  \centering
  
  \begin{tabular}{llllll}
    \toprule
    &               \multicolumn{5}{c}{\boldsymbol{$\mathcal{L}_{1}$}\textbf{ distance}} \\
                    \cmidrule(r){2-6} 
    \textbf{Method}                 &\textbf{Adult}         &\textbf{Breast cancer}     &\textbf{Covertype}         &\textbf{Portuguese Bank}       &\textbf{Spambase}               \\
    \midrule
    LORE                            &$0.11\pm0.04$          &-          	            &$0.11\pm0.03$              &$0.12\pm0.03$                  &$0.09\pm0.03$    \\
    MO	                            &$0.39\pm0.07$          &-                          &$0.46\pm0.01$	            &$0.31\pm0.08$	                &$0.30\pm0.01$    \\
    DiCE (random)	                &$1.76\pm0.10$	        &-	                        &$0.44\pm0.04$	            &$1.39\pm0.21$	                &$0.28\pm0.05$    \\
    DiCE (genetic)	                &$0.09\pm0.02$	        &-	                        &$0.92\pm0.09$	            &$0.60\pm0.14$	                &$0.31\pm0.02$    \\
    DiCE (binary, random)               &$1.76\pm0.10$	        &-	                        &$0.44\pm0.04$	            &$1.39\pm0.21$	                &$0.28\pm0.05$    \\
    DiCE (binary, genetic)              &$0.08\pm0.02$	        &-	                        &$0.96\pm0.06$	            &$0.59\pm0.13$	                &$0.31\pm0.02$    \\
    Ours (w/o AE)                    &$0.23\pm0.07$	        &-  	                    &$0.57\pm0.14$	            &$0.18\pm0.09$	                &$0.26\pm0.09$      \\                   
    Ours (w/o AE, diverse)              &$0.24\pm0.05$	        &-	                        &$0.56\pm0.10$	            &$0.32\pm0.17$	                &$0.23\pm0.08$      \\
    Ours                            &$0.19\pm0.06$	        &-	                        &$0.32\pm0.11$	            &$0.15\pm0.11$	                &$0.17\pm0.02$    \\
    Ours (diverse)                      &$0.21\pm0.05$	        &-	                        &$0.32\pm0.07$	            &$0.24\pm0.13$	                &$0.15\pm0.02$    \\  

    \bottomrule
  \end{tabular}
\end{table*}

%% file: subsections/04-experiments/02-sparsity.tex
\subsection{Sparsity}
\label{subsec:experiments_sparsity}

We report two sparsity measures between the original input and a valid counterfactual: $\mathcal{L}_{0}$ distance over categorical and $\mathcal{L}_{1}$ distance over standardized numerical features. Fair evaluation between methods that achieve different levels of validity is however not trivial. For example, reporting the sparsity over all valid counterfactuals per algorithm can favor methods that achieve a lower validity score. Finding a valid counterfactual instance may require larger displacements from the original input to reach the intended target, which will decrease the sparsity for the methods which managed to succeed and will not affect the ones that failed. On the other hand, performing a pairwise comparison over the intersection of valid counterfactuals can have a similar effect, favoring a low validity method that follows a policy focused on changing a single feature value.  In Tables \ref{tab:l0} and \ref{tab:l1}, we report the $\mathcal{L}_{0}$ and $\mathcal{L}_{1}$ metrics across the valid counterfactuals to confirm the sparsity of our results. Most comparable to our proposal are the MO and DiCE (random) approaches which achieve similar validity over four datasets (Adult, Breast Cancer, Portuguese Bank, Spambase). Compared to MO, our method generates significantly sparser counterfactuals across all datasets for both numerical and categorical features. DiCE returns  sparser counterfactuals on categorical features but does significantly worse on the numerical ones compared to our method which manages to balance the $\mathcal{L}_{0}$ and $\mathcal{L}_{1}$ losses successfully.

%% file: subsections/04-experiments/039-umap_figure.tex
\begin{figure}[!htbp]
    \centering
    \includegraphics[width=\linewidth]{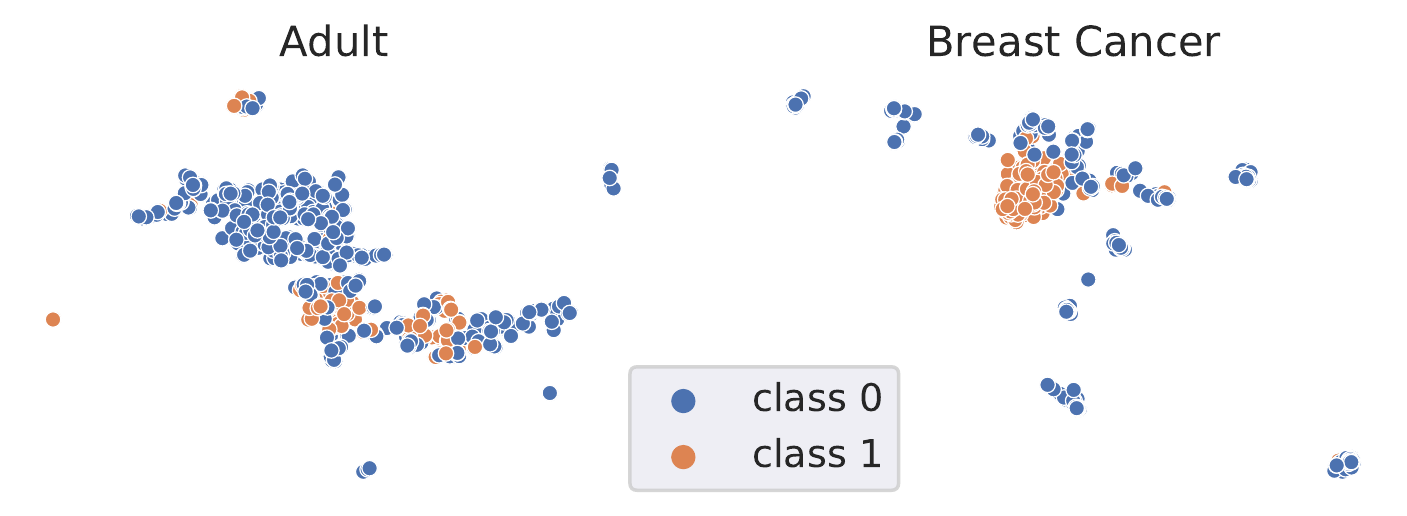}
    \caption{UMAP \cite{mcinnes2018uniform} embeddings of the Adult and Breast Cancer training sets, labeled by a logistic regression model.}
    \label{fig:umap}
\end{figure}

%% file: subsections/04-experiments/029-in_distribution_tables.tex
\begin{table*}[!htbp]
  \small
  \caption{In-distributionness---negative class (0) conditional MMD (mean$\pm$std over 5 classifiers and 3 random seeds).}     
  \label{tab:mmd0}
  \centering

  \begin{tabular}{llllll}
    \toprule
    &               \multicolumn{5}{c}{\boldsymbol{$MMD^2_{0}$}$(10^{-1})$} \\
                    \cmidrule(r){2-6} 
    \textbf{Method}                 &\textbf{Adult}                 &\textbf{Breast cancer}     &\textbf{Covertype}         &\textbf{Portuguese Bank}       &\textbf{Spambase}      \\
    \midrule
    LORE                            &$0.31\pm0.06$	    &$1.09\pm0.16$	    &$0.08\pm0.03$	    &$0.08\pm0.03$	    &$0.26\pm0.02$ \\
    MO	                            &$0.45\pm0.04$	    &$0.50\pm0.14$	    &$0.07\pm0.06$	    &$0.16\pm0.10$	    &$0.61\pm0.11$ \\
    DiCE (random)	                &$0.56\pm0.44$	    &$1.03\pm0.10$	    &$0.17\pm0.06$	    &$1.75\pm0.71$	    &$0.80\pm0.19$ \\
    DiCE (genetic)	                &$0.28\pm0.03$	    &$0.85\pm0.09$	    &$0.24\pm0.10$	    &$0.68\pm0.17$	    &$0.25\pm0.01$ \\
    DiCE (binary, random)           &$0.56\pm0.44$	    &$1.03\pm0.10$	    &$0.17\pm0.06$	    &$1.75\pm0.71$	    &$0.80\pm0.19$ \\
    DiCE (binary, genetic)          &$0.25\pm0.02$	    &$0.78\pm0.12$	    &$0.27\pm0.10$	    &$0.67\pm0.16$	    &$0.25\pm0.02$ \\
    Ours (w/o AE)                   &$0.56\pm0.31$	    &$1.01\pm0.41$	    &$0.33\pm0.40$	    &$0.07\pm0.05$	    &$0.39\pm0.53$ \\
    Ours (w/o AE, diverse)          &$0.45\pm0.22$	    &$1.02\pm0.18$	    &$0.25\pm0.20$	    &$0.05\pm0.03$	    &$0.27\pm0.40$ \\
    Ours                            &$0.36\pm0.06$	    &$0.36\pm0.34$	    &$0.04\pm0.02$	    &$0.10\pm0.07$	    &$0.32\pm0.13$ \\
    Ours (diverse)                  &$0.30\pm0.06$	    &$0.65\pm0.15$	    &$0.02\pm0.01$	    &$0.07\pm0.04$	    &$0.28\pm0.11$ \\
    \bottomrule
  \end{tabular}
\end{table*}

\begin{table*}[!htbp]
  \small
  \caption{In-distributionness---positive class (1) conditional MMD (mean$\pm$std over 5 classifiers and 3 random seeds).}      
  \label{tab:mmd1}
  \centering
  
  \begin{tabular}{llllll}
    \toprule
    &               \multicolumn{5}{c}{\boldsymbol{$MMD^2_{1}$}$(10^{-1})$} \\
                    \cmidrule(r){2-6} 
    \textbf{Method}                 &\textbf{Adult}                 &\textbf{Breast cancer}     &\textbf{Covertype}         &\textbf{Portuguese Bank}       &\textbf{Spambase}               \\
    \midrule
    LORE                            &$0.29\pm0.02$      &$0.80\pm0.35$	    &$0.07\pm0.08$	    &$0.19\pm0.04$	    &$0.24\pm0.05$ \\
    MO	                            &$0.13\pm0.01$	    &$1.53\pm0.62$	    &$0.10\pm0.06$	    &$0.16\pm0.09$	    &$0.76\pm0.19$ \\
    DiCE (random)	                &$1.48\pm0.27$	    &$0.46\pm0.10$	    &$0.15\pm0.07$	    &$1.50\pm0.38$	    &$0.75\pm0.17$ \\
    DiCE (genetic)	                &$0.34\pm0.04$	    &$0.37\pm0.20$	    &$0.07\pm0.07$	    &$0.15\pm0.10$	    &$0.22\pm0.07$ \\
    DiCE (binary, random)           &$1.48\pm0.28$	    &$0.46\pm0.10$	    &$0.15\pm0.07$	    &$1.50\pm0.38$	    &$0.75\pm0.18$ \\
    DiCE (binary, genetic)          &$0.33\pm0.04$	    &$0.48\pm0.22$	    &$0.07\pm0.07$  	&$0.15\pm0.11$	    &$0.22\pm0.07$ \\
    Ours (w/o AE)                   &$0.34\pm0.06$	    &$1.09\pm0.42$	    &$0.33\pm0.18$	    &$0.20\pm0.15$   	&$0.67\pm0.50$ \\
    Ours (w/o AE, diverse)          &$0.32\pm0.06$	    &$0.50\pm0.30$	    &$0.15\pm0.10$	    &$0.14\pm0.06$	    &$0.35\pm0.22$ \\
    Ours                            &$0.43\pm0.13$	    &$0.97\pm0.50$	    &$0.12\pm0.05$	    &$0.31\pm0.21$	    &$0.33\pm0.18$ \\
    Ours (diverse)                  &$0.30\pm0.06$	    &$0.28\pm0.15$	    &$0.09\pm0.03$	    &$0.21\pm0.16$	    &$0.23\pm0.16$ \\
    \bottomrule
  \end{tabular}
\end{table*}

%% file: subsections/04-experiments/049-diversity_figure.tex
\begin{figure*}[!htbp]
    \centering
    \includegraphics[width=0.95\textwidth]{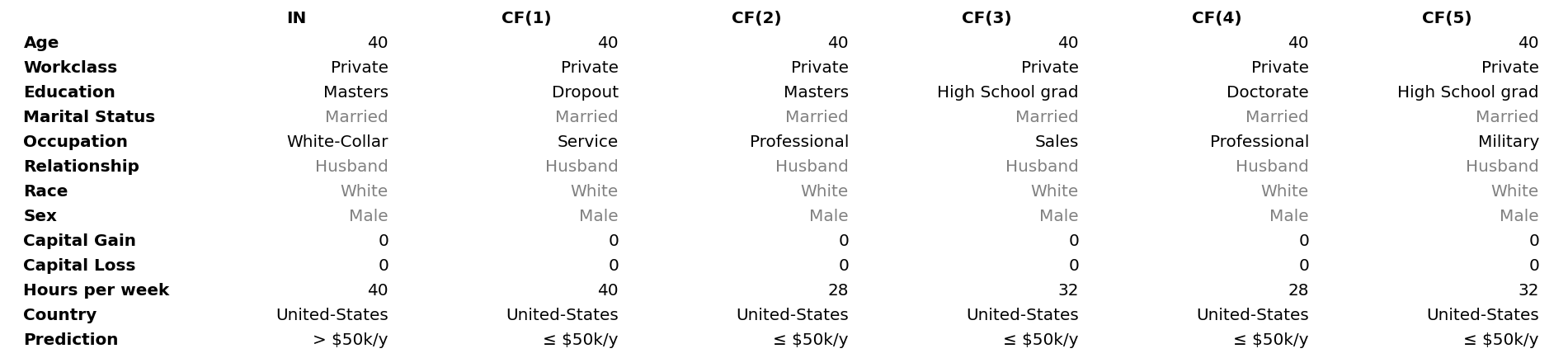}
    \caption{Diverse counterfactual instances via feature conditioning. IN---original instance, CF(i)---counterfactual instance. Grayed out feature values correspond to immutable features.}
    \label{fig:diversity}
\end{figure*}

%% file: subsections/04-experiments/051-flexibility_figure.tex
\begin{figure*}[!htbp]
    \centering
    \includegraphics[width=0.95\linewidth]{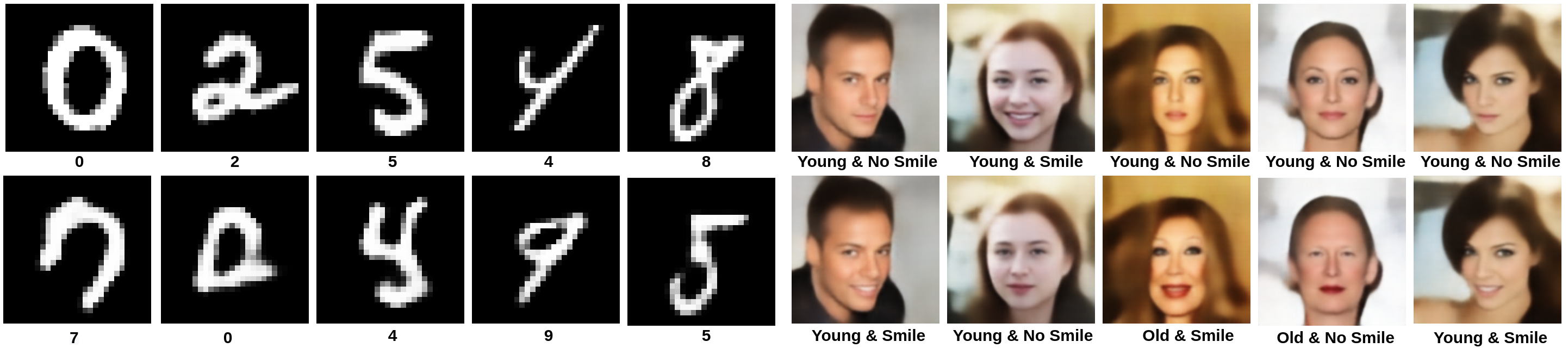}
    \caption{MNIST (left) and CelebA (right) counterfactual instances. On the top row, the original/reconstructed image for MNIST/CelebA respectively, and on the bottom, the counterfactual instance. We use autoencoder reconstructed images for CelebA to compare with the generated counterfactuals as the method is limited by the quality of the autoencoder. Higher quality reconstructions by the autoencoder would lead to higher quality counterfactuals. Labels under images are classifier predictions.}
    \label{fig:mnist_celeba}
\end{figure*}

%% file: subsections/04-experiments/03-in_distribution.tex
\subsection{In-distributionness}
\label{subsec:experiments_indistribution}

To evaluate the realism of the generated counterfactuals we compute the target-conditional MMD and compare it across the different methods in Tables \ref{tab:mmd0} and \ref{tab:mmd1}. The MMD is conditioned on the target prediction because we want the counterfactual to resemble feasible, in-distribution instances from the target class. See Appendix~\ref{app:in_distributioness} for MMD computational details.

It is not straightforward to fairly compare different methods given the difference in counterfactual validity. If we only consider the valid counterfactual instances, the comparison would be biased and is likely to favor methods with low validity which generate few but sparse, easy to flip counterfactuals. Even if we consider both the valid and invalid instances, the comparison depends on how overlapping the class-conditional distributions are. If they are clearly separated, the MMD provides a fair metric for direct comparison, but if the distributions overlap then the invalid instances might end up lowering the MMD which is undesirable. This is illustrated in Figure \ref{fig:umap}, which visualizes UMAP \cite{mcinnes2018uniform} embeddings of the Adult and Breast Cancer datasets, labeled by a logistic regression classifier. There is considerable overlap between the distributions of the two classes in Adult, leading to lower (better) MMD scores for methods with low validity such as LORE compared to our method. The classes in the Breast Cancer dataset are, however, more clearly separated, resulting in higher (worse) MMD values for LORE due to the invalid counterfactuals. As a result, the conditional MMD as a measure of in-distributionness should not be judged in isolation, but jointly with the validity and sparsity metrics.


DiCE (random) achieves similar validity to our method but reaches a higher (worse) MMD score on three datasets (Adult, Portuguese Bank, Spambase) and performs similarly on the other two. MO actually performs worse on two (Breast Cancer, Spambase) and similarly on the other three datasets compared to ours. This is a strong result for our method since the counterfactuals obtained by MO are actual instances of the target class from the training set.


%% file: subsections/04-experiments/04-diversity.tex
\subsection{Diversity}
\label{subsec:experiments_diversity}

Here we demonstrate that our method can be extended to generate diverse counterfactuals. Note that the deterministic decoding phase ensures consistency over repeated queries but limits the output to a single possible counterfactual per instance. To increase the diversity, we can sample the conditional vector subject to the user-defined feature constraints.  Thus, for unconstrained features, we follow the same sampling procedure applied during training, while for constrained ones, we sample a subset of their restricted values. The quantitative results for this setting show that such conditional generation produces in-distribution counterfactuals at the expense of a slight drop in validity performance (see row \emph{Ours (diverse)} in Tables~\ref{tab:counterfactuals_accuracy},\ref{tab:mmd0},\ref{tab:mmd1}). Figure \ref{fig:diversity} shows a sample of diverse counterfactuals for an instance in the Adult dataset.


%% file: subsections/04-experiments/05-flexibility.tex
\subsection{Flexibility to other data modalities}
\label{subsec:experiments_data_modaliites}
%

To demonstrate the flexibility of our approach, we ran experiments on two image datasets, MNIST---a collection of handwritten digits~\cite{lecun-mnisthandwrittendigit-2010}, and CelebA---a collection of face images~\cite{liu2015faceattributes} divided into four non-overlapping classes characterized by smiling/non-smiling and young/old classification labels. The classification model (CNN) is the same as in \cite{van2019interpretable, van2021conditional}, achieving accuracy on the test set of 99\% and 81\% for MNIST and CelebA respectively. Our training pipeline remains unchanged and only requires a pre-trained autoencoder for each dataset (see Appendix~\ref{app:img_ae}).
We trained the MNIST and CelebA counterfactual generator for 300K and 100K iterations, with a sparsity coefficient $\lambda_s=7.5$ for MNIST, and $\lambda_s=20.0$ for CelebA, and a consistency coefficient $\lambda_c=0$ for both datasets.
In Figure \ref{fig:mnist_celeba} we show samples of counterfactual instances for both datasets which demonstrates that our method is able to generate valid, in-distribution counterfactuals for high-dimensional data. See Appendices~\ref{app:mnist} and \ref{app:celeba} for more samples.

%% file: sections/06-conclusion.tex
\section{Conclusion}
\label{sec:conclusion}

Counterfactual instances are a powerful method for explaining the predictions of automated, black-box decision-making systems. They can be particularly useful for actionable recourse by providing direct feedback of what changes to the individual features could be made in order to achieve a specified outcome. Furthermore, being able to condition the explanations on a desired target as well as feasible feature ranges ensures that only relevant explanations are generated.

In this paper, we propose a model-agnostic, conditional counterfactual generation method. We focused primarily on the tabular data setting, and showed that our approach can generate batches of sparse, in-distribution counterfactual instances across various datasets, allowing for target and feature conditioning. Our method is stable during training and does not require extensive hyperparameter tuning. Additionally, we demonstrated that our method is flexible and extendable with minimal effort to other data modalities.

%% file: sm_sections/00-tabular_classifier.tex
\section{Tabular classifiers}\label{app:tab_class}

We evaluated our method on five classifiers. Logistic Regression (LR), Decision Tree (DT), Random Forest (RF), Linear Support Vector Classifier (LSVC) are part of the Scikit-learn machine learning library, and XGBoost Classifier is part of the XGBoost library. Table \ref{tab:classifiers_hyperparameters} presents the cross-validation hyperparameters. Missing values correspond to the default ones. We preprocessed each dataset by standardizing the numerical variables and one-hot encoding the categorical ones.

\begin{table*}[!htbp]
  \small
  \caption{Tabular classifiers' hyperparameters settings.}
  \label{tab:classifiers_hyperparameters}
  \centering
  
  \begin{tabular}{p{2cm}p{2cm}p{2cm}p{2cm}p{2cm}p{2cm}}
    \toprule
    &               \multicolumn{5}{c}{\textbf{Datasets}} \\
                    \cmidrule(r){2-6} 
    \textbf{Model}       &\textbf{Adult}    &\textbf{Breast cancer}     &\textbf{Covertype}   &\textbf{Portuguese Bank}   &\textbf{Spambase}      \\
    \midrule
    LR                   &C: 10     &C: 0.1     &C: 100     &C: 10      &C: 0.1      \\
    \midrule
    DT	                 &max\_depth: 10, min\_samples\_split: 5    &max\_depth: 3, min\_samples\_split: 4      &max\_depth: 25, min\_samples\_split: 3                                              &max\_depth: 25                            &max\_depth: 10, min\_samples\_split: 3           \\
    \midrule
    RF          	     &max\_depth: 15, min\_samples\_split: 10, n\_estimators: 50  &max\_depth: 4, min\_samples\_split: 4, n\_estimators: 50  &max\_depth: 25, n\_estimators: 50                   &max\_depth: 25, min\_samples\_split: 3, n\_estimators: 50   &max\_depth: 20, min\_samples\_split: 4, n\_estimators: 50           \\
    \midrule
    LSCV        	     &C: 1  &C=0.01     &C: 10  &C: 100     &C=0.01          \\
    \midrule
    XGBC                &min\_child\_weight: 0.5, max\_depth: 3, gamma: 0.2     &min\_child\_weight: 0.1, max\_depth: 3, gamma: 0   &min\_child\_weight: 0.1, max\_depth: 20, gamma: 0            &min\_child\_weight: 5.0, max\_depth: 4, gamma: 0.01    &min\_child\_weight: 1.0, max\_depth: 20, gamma: 0.1            \\
    \bottomrule
  \end{tabular}
\end{table*}

%% file: sm_sections/01-tabular_autoencoder.tex
\section{Tabular autoencoders}\label{app:tab_ae}

All tabular autoencoders are fully connected networks with a multi-headed decoder to properly address heterogeneous datatypes. We modeled numerical features using normal distributions with constant variance and categorical ones using categorical distributions. Table \ref{tab:tabular_autoenc_arch} summarizes the network architectures of the autoencoder models for each dataset. We trained all models for 100K steps using Adam optimizer~\cite{kingma2014adam} with a batch size of 128 and a learning rate of 1e-3. We defined the reconstruction loss as a weighted combination between Squared Error (SE) averaged across all numerical features and Cross-Entropy (CE) averaged across the categorical ones. In all our experiments, we consider a weight equal to 1 for both loss terms.

\begin{table*}[!htbp]
  \caption{Tabular autoencoders' architectures.}
  \label{tab:tabular_autoenc_arch}
  \centering
  
  \small
  \begin{tabular}{lllll}
    \toprule
    \multicolumn{5}{c}{\textbf{Datasets}} \\
    \cmidrule(r){0-4} 
    \textbf{Adult}         &\textbf{Breast cancer}     &\textbf{Covertype}     &\textbf{Portuguese Bank}     &\textbf{Spambase}\\
    \midrule
    Input Layer            &Input Layer                &Input Layer            &Input Layer                  &Input Layer\\
    Linear(128)            &Linear(70)                 &Linear(128)            &Linear(128)                  &Linear(128) \\
    ReLU()                 &Tanh()                     &ReLU()                 &ReLU()                       &ReLU() \\
    Linear(15)             &Output Layer               &Liner(15)              &Linear(15)                   &Linear(50) \\ 
    Tanh()                 &-                          &Tanh()                 &Tanh()                       &Tanh() \\
    Linear(128)            &-                          &Linear(128)            &Linear(128)                  &Linear(128) \\
    ReLU()                 &-                          &ReLU()                 &ReLU()                       &ReLU() \\
    Output Layer           &-                          &Output Layer           &Output Layer                 &Output Layer\\
    \bottomrule
  \end{tabular}
\end{table*}

%% file: sm_sections/02-image_autoencoder.tex
\section{Image autoencoders}\label{app:img_ae}

The MNIST encoder projects the input instances to a latent representation of dimension 32. We trained the autoencoder for 50K steps with a batch size of 64, optimizing the Binary Cross-Entropy (BCE) using Adam optimizer~\cite{kingma2014adam} with a learning rate of 1e-3. The CelebA encoder projects the input space from 128x128 to a latent representation of dimension 128. We trained the autoencoder for 350K steps with a batch size of 128, optimizing the Mean Squared Error(MSE) using Adam optimizer~\cite{kingma2014adam} with a learning rate of 1e-4.  Tables \ref{tab:image_autoenc_arch} and \ref{tab:image_blocks_arch} summarize the autoencoders' architecture for both datasets.

\begin{table*}[!htbp]
  \caption{Image autoencoders' architectures.}
  \label{tab:image_autoenc_arch}
  \centering
  
  \begin{tabular}{ll}
    \toprule
    \multicolumn{2}{c}{\textbf{Datasets}} \\
    \cmidrule(r){0-1} 
    \textbf{MNIST}                                          &\textbf{CelebA}\\
    \midrule
    Conv2d(1, 16, kernel\_size=(3, 3), padding=1)           &ConvBlock(3, 32)\\
    ReLU()                                                  &ConvBlock(32, 64)\\
    MaxPool2d(kernel\_size=(2, 2), stride=2)                &ConvBlock(64, 128\\
    Conv2d(16, 8, kernel\_size=(3, 3), padding=1)           &ConvBlock(128, 256)\\
    ReLU()                                                  &ConvBlock(256, 512)\\
    MaxPool2d(kernel\_size=(2, 2), stride=2)                &ConvBlock(512, 512)\\
    Conv2d(8, 8, kernel\_size=(3, 3), padding=2)            &Flatten()\\
    ReLU()                                                  &Linear(2048, 128)\\
    MaxPool2d(kernel\_size=(2, 2), stride=2)                &Tanh()\\
    Flatten()                                               &Reshape(512, 2, 2) \\
    Linear(128, 32)                                         &TransConvBlock(512, 512)\\
    Tanh()                                                  &TansConvBlock(512, 512) \\
    Linear(32, 128)                                         &TransConvBlock(512, 256)\\
    ReLU()                                                  &TransConvBlock(256, 128) \\
    Reshape(8, 4, 4)                                        &TransConvBlock(128, 64) \\
    Conv2d(8, 8, kernel\_size=(3, 3), padding=1)            &TransConvBlock(64, 32) \\
    ReLU()                                                  &TransConvBlock(32, 32) \\
    Upsample(scale\_factor=2)                               &TransConvBlock(32, 3) \\
    Conv2d(8, 8, kernel\_size=(3, 3), padding=1)            &Tanh()\\
    ReLU()                                                  &- \\
    Upsample(scale\_factor=2)                               &- \\
    Conv2d(8, 16, kernel\_size=(3, 3))                      &- \\
    ReLU()                                                  &- \\
    Upsample(scale\_factor=2)                               &- \\
    Conv2d(16, 1, kernel\_size=(3, 3), padding=1)           &- \\
    Sigmoid()                                               &- \\
    \bottomrule
  \end{tabular}
\end{table*}

\begin{table*}[!htbp]
  \small
  \caption{Blocks' architectures.}
  \label{tab:image_blocks_arch}
  \centering
  
  \begin{tabular}{p{6.5cm}p{6.5cm}}
    \toprule
    \multicolumn{2}{c}{\textbf{Blocks}} \\
    \cmidrule(r){0-1} 
    \textbf{ConvBlock}         &\textbf{TransConvBlock}\\
    \midrule
    Conv2d(in\_channels, out\_channels, kernel\_size=3, stride=2, padding=1)        &ConvTranspose2d(in\_channels, out\_channels, kernel\_size=3, stride=2, padding=1, output\_padding=1) \\
    BatchNorm2d(out\_channles)                                                      &BatchNorm2d(out\_channles) \\
    LeakyReLU()                                                                     &LeakyReLU() \\
    \bottomrule
  \end{tabular}
\end{table*}

%% file: sm_sections/03-ddpg.tex
\clearpage
\section{DDPG}\label{app:ddpg}

The actor and the critic networks follow the same architecture for both data modalities, as displayed in Table \ref{tab:ddpg_arch}. We trained the models for 100K steps using a batch size of 128 and Adam optimizer~\cite{kingma2014adam} with the same learning rate of 1e-3 for both actor and critic. We enforced exploration by uniformly sampling the embedding representation in the interval [-1, 1] for 100 training steps. Afterwards, we applied Gaussian additive noise with a constant standard deviation of 0.1 for the rest of the training. The replay buffer has a capacity of 1000 batches, equivalent to 128000 experiences. The learning process starts after ten iterations, corresponding to a replay buffer containing 1280 experiences.

\begin{table*}[!htbp]
  \small
  \caption{DDPG actor-critic architectures.}
  \label{tab:ddpg_arch}
  \centering
  
  \begin{tabular}{ll}
    \toprule
    \textbf{Actor}                           &\textbf{Critic}\\
    \midrule
    Linear(input\_size, hidden\_dim)         &Linear(input\_size + latent\_dim, hidden\_dim)\\
    LayerNorm(hidden\_dim)                   &LayerNorm(hidden\_dim)\\
    ReLU()                                   &ReLU()\\
    Linear(hidden\_dim, hidden\_dim)         &Linear(hidden\_dim, hidden\_dim)\\
    LayerNorm(hidden\_dim)                   &LayerNorm(hidden\_dim)\\
    ReLU()                                   &ReLU()\\
    Linear(hidden\_dim, latent\_dim)         &Linear(hidden\_dim, 1)\\
    \bottomrule
  \end{tabular}
\end{table*}

%% file: sm_sections/04-mmd_covertype.tex
\section{In-distributionness}\label{app:in_distributioness}

We evaluate the realism of the generated counterfactual through the target-conditional MMD. We condition the MMD on the target prediction since we want to measure the in-distributionness of the generated instances against the reference training instances belonging to the target class. We perform a dimensionality reduction step for the MMD computation with a randomly initialized encoder~\cite{rabanser2019loudly} following an architecture that consists of three linear layers of dimension 32, 16, and 5 with ReLU non-linearities in-between. We use a radial basis function kernel for which we derive the standard deviation from the input data~\cite{alibi-detect}. Table \ref{tab:mmd_covertype} summarizes the results obtained for the \emph{Covertype} dataset for all classes, initially presented in Section \ref{subsec:experiments_indistribution} only for the first two most dominant classes along with the rest of binary classification datasets. The scores are averaged across three seeds and all classifiers.

\begin{table*}[!htbp]
  \small
  \caption{Covertype MMD for unbalanced datasets with untrained encoder dimensionality reduction}      
  \label{tab:mmd_covertype}
  \centering
  
  \begin{tabular}{llllllll}
    \toprule
    &               \multicolumn{7}{c}{\textbf{Target conditional MMD} ($10^{-1}$)} \\
                    \cmidrule(r){2-8} 
    \textbf{Method}                 &\textbf{0}       &\textbf{1}       &\textbf{2}         &\textbf{3}         &\textbf{4}         &\textbf{5}         &\textbf{6}   \\
    \midrule
    LORE                            &$0.08\pm0.03$    &$0.07\pm0.08$      &$0.52\pm0.07$    &$1.25\pm0.14$      &$0.23\pm0.11$      &$1.20\pm0.17$      &$0.25\pm0.04$ \\
    MO	                            &$0.07\pm0.06$	  &$0.10\pm0.06$	  &$0.47\pm0.09$    &$0.05\pm0.10$      &$0.05\pm0.04$      &$0.85\pm0.09$	    &$0.15\pm0.05$ \\
    DICE(random)	                &$0.17\pm0.06$    &$0.15\pm0.07$      &$0.94\pm0.72$    &$0.65\pm0.51$      &$0.63\pm0.47$      &$0.87\pm0.18$	    &$0.36\pm0.13$ \\
    DICE(genetic)	                &$0.24\pm0.10$    &$0.07\pm0.07$      &$0.41\pm0.28$    &$0.59\pm0.28$      &$0.38\pm0.17$      &$0.71\pm0.34$	    &$0.44\pm0.20$ \\
    DiCE(binary, random)            &$0.17\pm0.06$    &$0.15\pm0.07$      &$0.94\pm0.72$    &$0.65\pm0.51$      &$0.63\pm0.47$      &$0.87\pm0.18$      &$0.36\pm0.13$ \\
    DiCE(binary, genetic)           &$0.27\pm0.10$	  &$0.07\pm0.07$      &$0.47\pm0.26$    &$0.54\pm0.27$      &$0.36\pm0.18$      &$0.72\pm0.33$      &$0.46\pm0.20$ \\
    Ours(w/o AE)                    &$0.33\pm0.40$    &$0.33\pm0.18$      &$0.65\pm0.25$    &$1.04\pm0.51$      &$0.82\pm0.58$      &$0.78\pm0.31$      &$0.57\pm0.41$ \\
    Ours(w/o AE, diverse)           &$0.25\pm0.20$    &$0.15\pm0.10$      &$0.49\pm0.26$    &$0.75\pm0.52$      &$0.37\pm0.18$      &$0.52\pm0.34$      &$0.23\pm0.13$ \\
    Ours                            &$0.04\pm0.02$    &$0.12\pm0.05$      &$0.48\pm0.21$    &$0.55\pm0.46$      &$0.33\pm0.26$      &$0.78\pm0.48$      &$0.16\pm0.14$ \\
    Ours(diverse)                   &$0.02\pm0.01$    &$0.09\pm0.03$      &$0.22\pm0.08$    &$0.66\pm0.47$      &$0.33\pm0.22$	    &$0.60\pm0.30$      &$0.14\pm0.08$ \\
    \bottomrule
  \end{tabular}
\end{table*}

%% file: sm_sections/05-hyperparameters.tex
\clearpage
\section{Results accross different hyperparameters averaged over all datasets, classifiers and seeds.}\label{app:hyper}

We experiment with a secondary training pipeline by removing the autoencoder component. To map the input features to the interval $[-1, 1]$, we replace the encoding step with a projection through a \emph{tanh} nonlinearity and the decoding phase with its corresponding inverse transformation. The rest of the training procedure remains identical. 

For both pipeline configurations, we perform a hyperparameter search over the sparsity ($\lambda_s$) and consistency ($\lambda_c$) coefficients. We evaluate the percentage of the valid counterfactuals and the sparsity through the $\mathcal{L}_{0}$ and $\mathcal{L}_{1}$ norm for $1000$ instances, where the test set permits.  We report the results for the original unbalanced class distribution (denoted as unbalanced) and the results obtained after balancing the test set (denoted as balanced). The balancing procedure consists of sampling uniformly without replacement the same number of instances from the class conditional subset for each class. Table~\ref{tab:comparison} summarizes the results averaged across all seeds, classifiers, and datasets. Note that sparser results correspond to lower validity scores. Additional per dataset, per classifier hyperparameter selection can yield better results.


\begin{table*}[!htbp]
  \footnotesize
  \caption{Comparison validity and sparsity}
  \label{tab:comparison}
  \centering
  
  \begin{tabular}{lllllllll}
    \toprule
    &\multicolumn{3}{c}{\textbf{Unbalanced}}    &\multicolumn{3}{c}{\textbf{Balanced}} \\
                    \cmidrule(r){2-4}   \cmidrule{5-7}
    \textbf{Method}                 &\textbf{Validity}(\%)      &\boldsymbol{$\mathcal{L}_{0}$}         &\boldsymbol{$\mathcal{L}_{1}$}   &\textbf{Validity}(\%)      &\boldsymbol{$\mathcal{L}_{0}$}         &\boldsymbol{$\mathcal{L}_{1}$}\\
    \midrule
    $\lambda_s=0.1$, w/o AE                   &$96.01\pm6.69$	    &$0.35\pm0.17$	    &$0.31\pm0.18$	&$96.09\pm5.97$	    &$0.34\pm0.17$	&$0.31\pm0.17$	\\
    $\lambda_s=0.2$, w/o AE                   &$93.89\pm8.44$	    &$0.28\pm0.14$	    &$0.25\pm0.14$	&$93.89\pm7.76$	    &$0.28\pm0.14$	&$0.25\pm0.13$  \\
    $\lambda_s=0.3$, w/o AE                   &$90.94\pm12.24$	    &$0.25\pm0.12$	    &$0.23\pm0.13$	&$90.47\pm11.50$	&$0.24\pm0.13$	&$0.23\pm0.11$  \\
    $\lambda_s=0.4$, w/o AE                   &$88.07\pm14.75$	    &$0.24\pm0.12$	    &$0.22\pm0.12$	&$86.92\pm14.84$	&$0.23\pm0.12$	&$0.22\pm0.11$  \\
    $\lambda_s=0.5$, w/o AE                   &$84.63\pm18.67$	    &$0.22\pm0.11$	    &$0.20\pm0.11$	&$82.02\pm19.58$	&$0.21\pm0.11$	&$0.20\pm0.09$  \\
    \midrule
    $\lambda_s=0.1$, w/o AE, diverse          &$96.01\pm6.69$	    &$0.35\pm0.17$	    &$0.31\pm0.18$	&$96.09\pm5.97$	    &$0.34\pm0.17$	&$0.31\pm0.17$ \\
    $\lambda_s=0.2$, w/o AE, diverse          &$93.89\pm8.44$	    &$0.28\pm0.14$	    &$0.25\pm0.14$	&$93.89\pm7.76$	    &$0.28\pm0.14$	&$0.25\pm0.13$ \\
    $\lambda_s=0.3$, w/o AE, diverse          &$90.94\pm12.24$	    &$0.25\pm0.12$	    &$0.23\pm0.13$	&$90.47\pm11.50$	&$0.24\pm0.13$	&$0.23\pm0.11$ \\
    $\lambda_s=0.4$, w/o AE, diverse          &$88.07\pm14.75$	    &$0.24\pm0.12$	    &$0.22\pm0.12$	&$86.92\pm14.84$	&$0.23\pm0.12$	&$0.22\pm0.11$ \\
    $\lambda_s=0.5$, w/o AE, diverse          &$84.63\pm18.67$	    &$0.22\pm0.11$	    &$0.20\pm0.11$	&$82.02\pm19.58$	&$0.21\pm0.11$	&$0.20\pm0.09$ \\
    \midrule
    $\lambda_s=0.1, \lambda_c=0.0$            &$99.54\pm1.15$	    &$0.51\pm0.22$	    &$0.37\pm0.20$	&$99.35\pm1.15$	    &$0.51\pm0.21$	&$0.37\pm0.20$ \\
    $\lambda_s=0.1, \lambda_c=0.5$            &$99.03\pm2.98$	    &$0.46\pm0.27$	    &$0.38\pm0.21$	&$98.78\pm3.17$	    &$0.48\pm0.28$	&$0.37\pm0.20$ \\
    \midrule
    $\lambda_s=0.2, \lambda_c=0.5$            &$99.02\pm2.60$	    &$0.39\pm0.23$	    &$0.29\pm0.16$	&$98.65\pm3.01$	    &$0.40\pm0.24$	&$0.28\pm0.16$ \\
    \midrule
    $\lambda_s=0.3, \lambda_c=0.5$            &$97.81\pm6.74$	    &$0.34\pm0.20$	    &$0.26\pm0.14$	&$97.42\pm6.66$	    &$0.35\pm0.22$	&$0.24\pm0.13$  \\
    $\lambda_s=0.3, \lambda_c=1.0$            &$97.80\pm6.45$	    &$0.37\pm0.23$	    &$0.25\pm0.13$  &$97.49\pm6.39$ 	&$0.38\pm0.24$	&$0.24\pm0.12$  \\
    \midrule
    $\lambda_s=0.4, \lambda_c=0.5$            &$97.23\pm6.38$	    &$0.32\pm0.19$	    &$0.23\pm0.11$	&$95.99\pm7.85$	    &$0.33\pm0.20$	&$0.21\pm0.10$	\\
    $\lambda_s=0.4, \lambda_c=1.0$            &$97.20\pm7.33$	    &$0.33\pm0.21$	    &$0.22\pm0.11$  &$95.95\pm8.43$	    &$0.34\pm0.23$	&$0.21\pm0.10$  \\
    \midrule
    $\lambda_s=0.5, \lambda_c=0.0$            &$96.51\pm9.73$	    &$0.39\pm0.16$  	&$0.25\pm0.13$	&$95.21\pm10.40$	&$0.39\pm0.18$	&$0.24\pm0.12$  \\
    $\lambda_s=0.5, \lambda_c=0.1$            &$96.13\pm9.55$	    &$0.30\pm0.16$	    &$0.23\pm0.11$	&$94.65\pm10.23$	&$0.31\pm0.17$	&$0.22\pm0.10$  \\
    $\lambda_s=0.5, \lambda_c=0.2$            &$96.52\pm7.89$	    &$0.29\pm0.16$	    &$0.22\pm0.11$	&$94.74\pm9.97$	    &$0.30\pm0.17$	&$0.20\pm0.10$  \\
    $\lambda_s=0.5, \lambda_c=0.5$            &$96.42\pm7.85$	    &$0.30\pm0.18$	    &$0.21\pm0.11$	&$94.75\pm10.07$	&$0.31\pm0.19$	&$0.19\pm0.09$  \\
    $\lambda_s=0.5, \lambda_c=1.0$            &$96.57\pm7.97$	    &$0.31\pm0.20$	    &$0.21\pm0.10$	&$94.89\pm10.18$	&$0.32\pm0.21$	&$0.20\pm0.09$  \\
    $\lambda_s=0.5, \lambda_c=1.5$            &$95.81\pm9.94$	    &$0.32\pm0.21$	    &$0.22\pm0.11$	&$93.90\pm12.01$	&$0.33\pm0.23$	&$0.20\pm0.09$	\\
    $\lambda_s=0.5, \lambda_c=5.0$            &$85.30\pm23.86$	    &$0.40\pm0.27$	    &$0.22\pm0.11$	&$86.31\pm19.06$	&$0.40\pm0.28$	&$0.20\pm0.10$	\\
    \midrule
    $\lambda_s=0.1, \lambda_c=0.0$, diverse   &$96.81\pm5.36$	    &$0.44\pm0.19$	    &$0.41\pm0.19$	&$96.25\pm5.72$	    &$0.44\pm0.20$	&$0.40\pm0.18$ \\
    $\lambda_s=0.1, \lambda_c=0.5$, diverse   &$96.64\pm6.42$	    &$0.40\pm0.23$	    &$0.40\pm0.18$	&$96.21\pm6.75$	    &$0.41\pm0.24$	&$0.38\pm0.17$ \\
    \midrule
    $\lambda_s=0.2, \lambda_c=0.5$, diverse  &$96.12\pm6.92$	    &$0.34\pm0.20$	    &$0.31\pm0.14$	&$95.57\pm7.35$	    &$0.34\pm0.21$	&$0.29\pm0.13$ \\
    \midrule
    $\lambda_s=0.3, \lambda_c=0.5$, diverse   &$95.30\pm9.06$	    &$0.30\pm0.18$	    &$0.27\pm0.12$	&$94.76\pm9.08$	    &$0.31\pm0.20$	&$0.26\pm0.11$ \\
    $\lambda_s=0.3, \lambda_c=1.0$, diverse   &$95.42\pm8.77$	    &$0.31\pm0.20$	    &$0.27\pm0.12$  &$94.84\pm8.66$	    &$0.33\pm0.22$	&$0.26\pm0.11$ \\	
    \midrule
    $\lambda_s=0.4, \lambda_c=0.5$, diverse   &$94.22\pm9.85$	    &$0.28\pm0.17$	    &$0.25\pm0.11$	&$93.25\pm10.71$	&$0.29\pm0.19$	&$0.23\pm0.09$ \\
    $\lambda_s=0.4, \lambda_c=1.0$, diverse   &$94.17\pm10.47$	    &$0.29\pm0.19$  	&$0.25\pm0.11$  &$93.35\pm11.01$	&$0.29\pm0.20$	&$0.23\pm0.09$ \\
    \midrule
    $\lambda_s=0.5, \lambda_c=0.0$, diverse   &$93.56\pm11.62$	    &$0.34\pm0.15$	    &$0.28\pm0.10$	&$92.58\pm12.04$	&$0.34\pm0.17$	&$0.26\pm0.10$ \\
    $\lambda_s=0.5, \lambda_c=0.1$, diverse   &$93.29\pm11.90$	    &$0.27\pm0.15$	    &$0.25\pm0.10$	&$92.36\pm12.40$	&$0.28\pm0.17$	&$0.23\pm0.09$ \\
    $\lambda_s=0.5, \lambda_c=0.2$, diverse   &$93.81\pm10.81$	    &$0.26\pm0.15$	    &$0.24\pm0.10$	&$92.69\pm12.19$	&$0.27\pm0.17$	&$0.22\pm0.09$ \\
    $\lambda_s=0.5, \lambda_c=0.5$, diverse   &$93.14\pm11.29$	    &$0.26\pm0.16$	    &$0.23\pm0.10$	&$92.06\pm12.66$	&$0.27\pm0.18$	&$0.21\pm0.09$ \\
    $\lambda_s=0.5, \lambda_c=1.0$, diverse   &$93.40\pm11.45$	    &$0.27\pm0.17$	    &$0.23\pm0.10$	&$92.32\pm12.57$	&$0.28\pm0.19$	&$0.21\pm0.08$ \\
    $\lambda_s=0.5, \lambda_c=1.5$, diverse   &$92.85\pm12.18$  	&$0.27\pm0.18$	    &$0.23\pm0.10$	&$91.71\pm13.51$	&$0.28\pm0.20$	&$0.22\pm0.08$ \\
    $\lambda_s=0.5, \lambda_c=5.0$, diverse   &$81.39\pm24.63$	    &$0.33\pm0.24$	    &$0.24\pm0.10$	&$83.08\pm20.38$	&$0.34\pm0.26$	&$0.22\pm0.09$ \\
    \bottomrule
  \end{tabular}
\end{table*}

%% file: sm_sections/06-samples_adult.tex
\clearpage
\section{Samples Adult}\label{app:adult}
\begin{figure*}[!htbp]
    \centering
    \includegraphics[width=\textwidth]{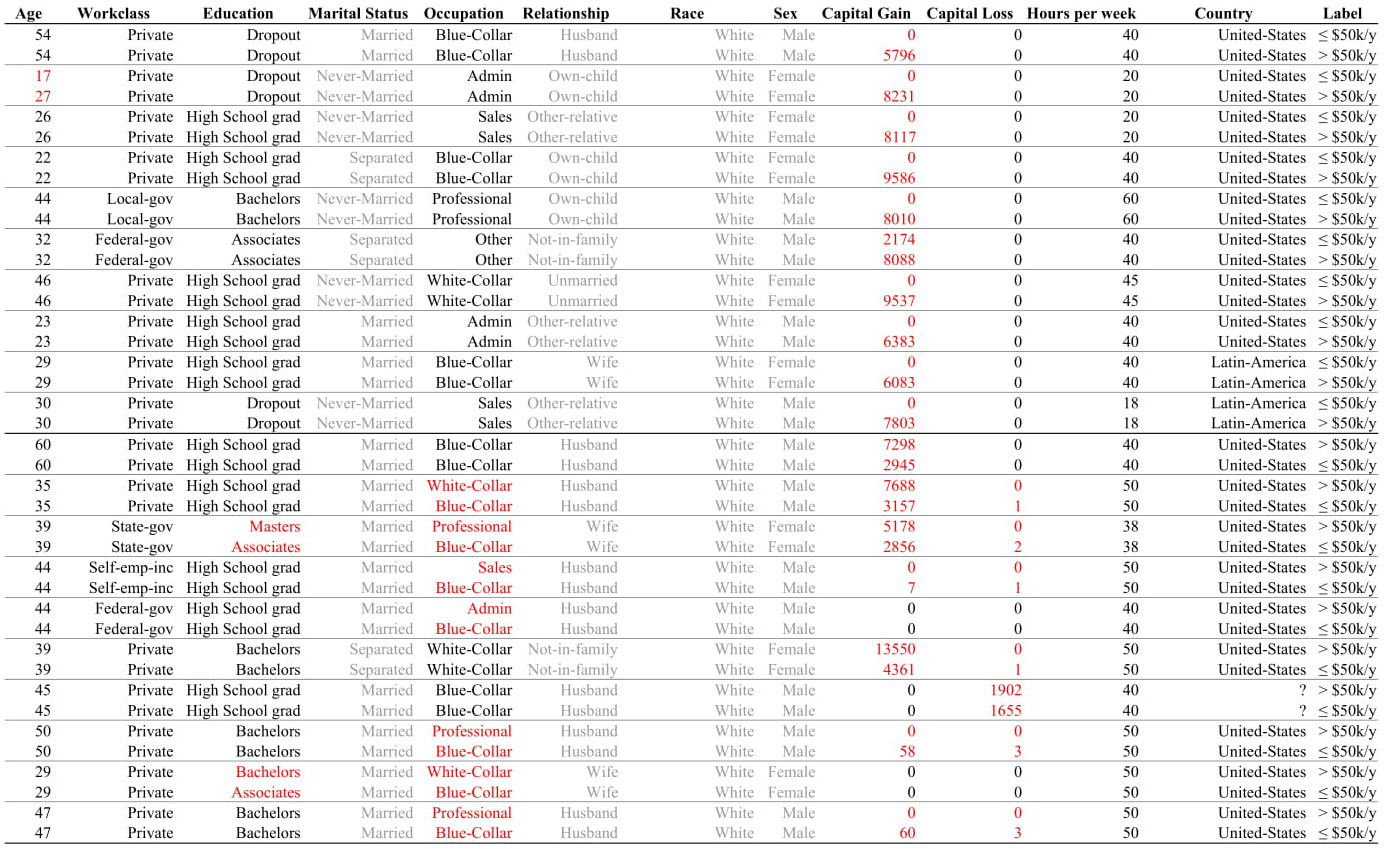}
    \caption{Counterfactual instances on Adult dataset. Odd rows correspond to the input instances and even rows correspond to the counterfactual instances. Grayed out features correspond to immutable features and \emph{Age} feature is allowed to increase only. Feature changes are highlighted in red.}
    \label{fig:adult_uncond}
\end{figure*}

\begin{figure*}[!htbp]
    \centering
    \includegraphics[width=\textwidth]{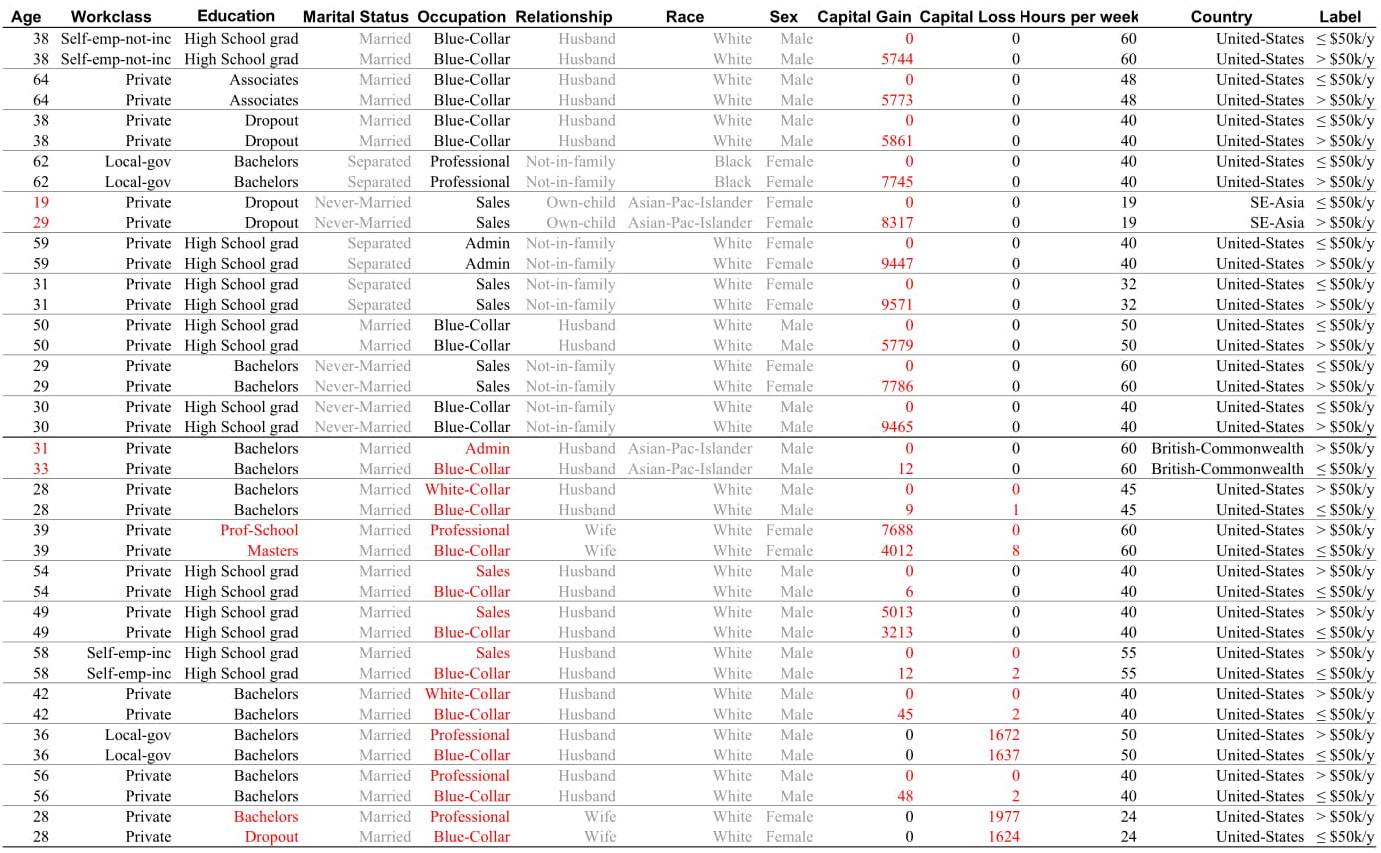}
    \caption{Conditional counterfactual instances on the Adult dataset. Odd rows correspond to the input instances and even rows correspond to the counterfactual instances. Grayed out features correspond to immutable features. All instances are conditioned on the same conditional vector corresponding to: 1) \emph{Age} allowed to increase by up to 10; 2) \emph{Workclass} allowed to change to \{\emph{Private}, \emph{Without-pay}\} or stay the same; 3) \emph{Education} allowed to change to \{\emph{Bachelors}, \emph{Masters}, \emph{Dropout}\} or stay the same; 4) \emph{Occupation} allowed to change to \{\emph{Sales}, \emph{White-Collar}, \emph{Blue-Collar}\} or stay the same; 5) \emph{Capital Gain} allowed to increase or decrease by up to 10000; 6) \emph{Capital Loss} allowed to increase or decrease by up to 10000; 7) \emph{Hours per week} allowed to increase or decrease by up to 20; 8) \emph{Country} cannot change. Feature changes are highlighted in red.}
    \label{fig:adult_cond}
\end{figure*}

%% file: sm_sections/07-samples_mnist.tex
\clearpage
\section{Samples MNIST}\label{app:mnist}

\begin{figure*}[!htbp]
    \centering
    \includegraphics[height=0.85\textheight, width=\textwidth]{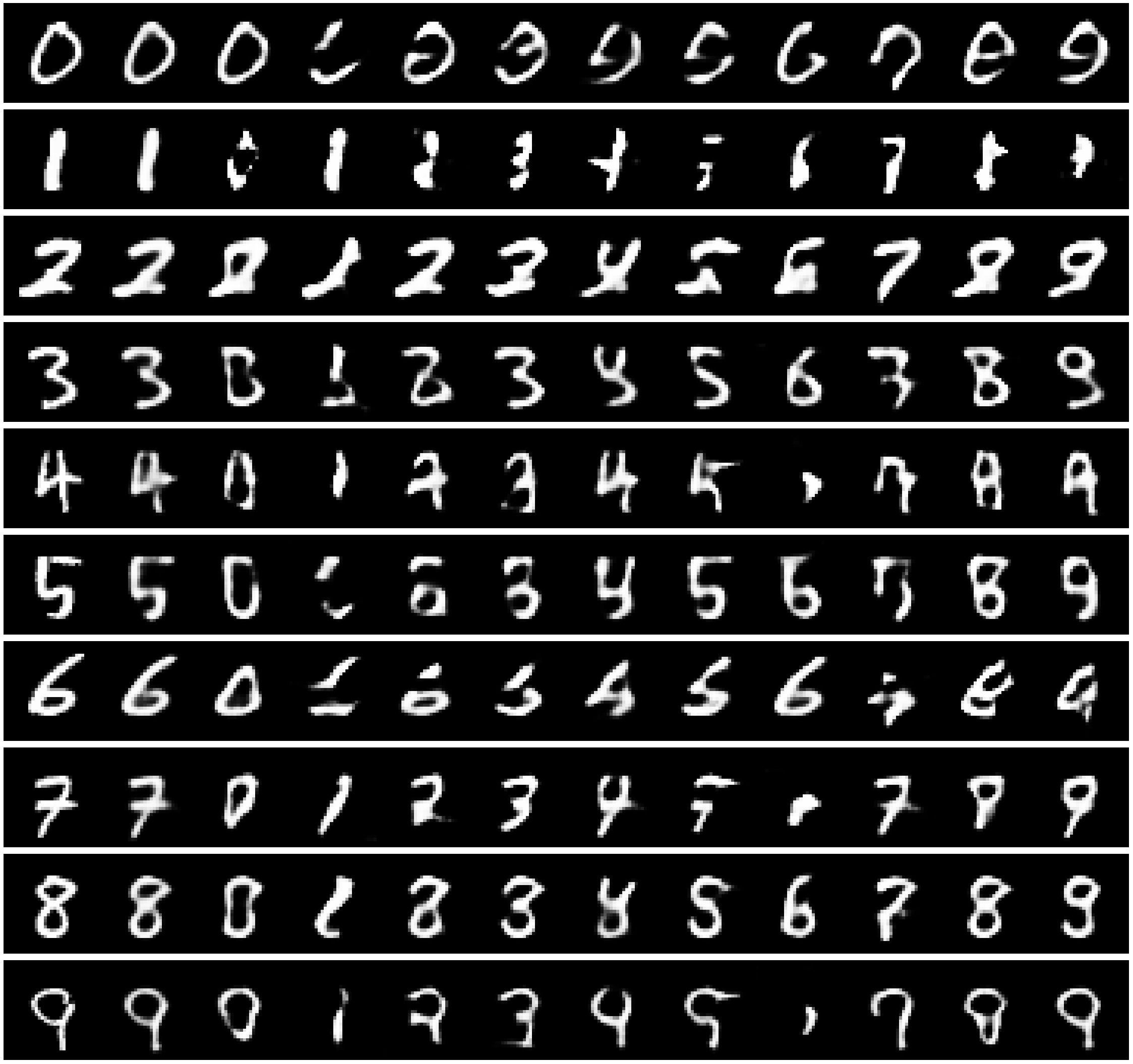}
    \caption{One instance to all classes on MNIST dataset. Starting from left to right: original instance, reconstructed instance, counterfactual instance for each class staring from 0 to 9.}
    \label{fig:mnist_samples_sm}
\end{figure*}

%% file: sm_sections/08-samples_celeba.tex
\clearpage
\section{Samples CELEBA}\label{app:celeba}

\begin{figure*}[!htbp]
    \centering
    \includegraphics[height=0.85\textheight]{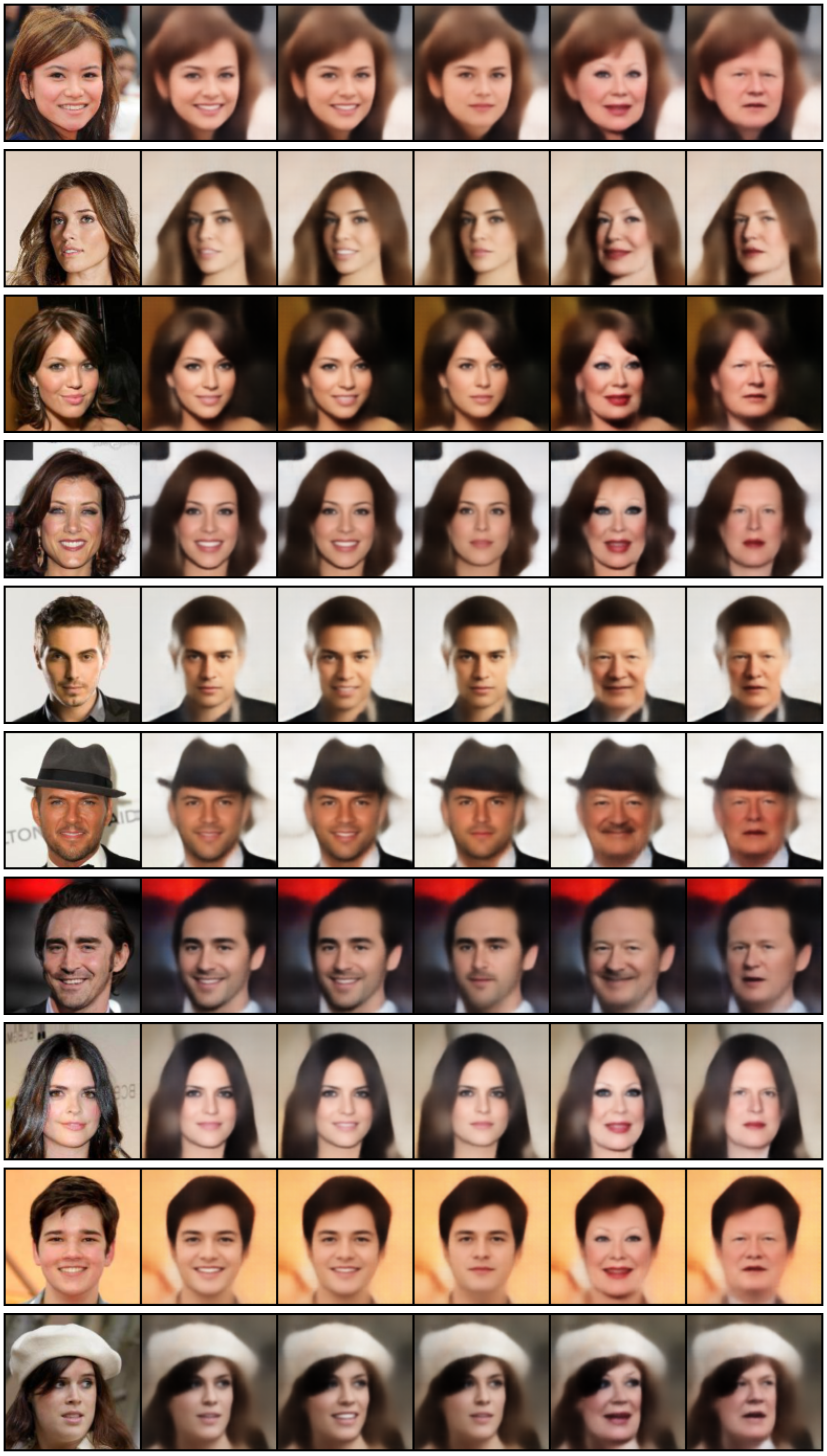}
    \caption{One instance to all classes on CelebA dataset. Starting from left to right: original instance, reconstructed instance, counterfactual instance for each class: young and smiling, young and not smiling, old and smiling, old and not smiling.}
    \label{fig:mnist_samples}
\end{figure*}